\documentclass[10pt,journal,compsoc]{IEEEtran}

\ifCLASSOPTIONcompsoc
  \usepackage[nocompress]{cite}
\else
  \usepackage{cite}
\fi

\usepackage{amsthm}
\usepackage{amssymb}
\usepackage{amsmath}
\usepackage{algorithm}
\usepackage{algorithmic}
\usepackage{graphicx}
\usepackage{hyperref}
\usepackage{enumerate}
\usepackage{xcolor}
\usepackage{booktabs}
\usepackage{subfig}

\ifCLASSINFOpdf
\else
\fi

\begin{document}

\title{Local Gaussian Processes for Efficient Fine-Grained Traffic Speed Prediction}

\author{Truc Viet Le,
        Richard Oentaryo,
        Siyuan Liu,
        and Hoong Chuin Lau
\IEEEcompsocitemizethanks{\IEEEcompsocthanksitem Truc Viet Le, Richard Oentaryo, and Hoong Chuin Lau are with the School of Information Systems, Singapore Management University, Singapore 178902. E-mail: \{trucviet.le.2012, roentaryo, hclau\}@smu.edu.sg.\protect\\
\IEEEcompsocthanksitem Siyuan Liu is with the Smeal College of Business, Pennsylvania State University, University Park, PA 16802. E-mail: siyuan@psu.edu.}
\thanks{Manuscript received on November 30, 2015; revised on July 30, 2016.}}


\IEEEtitleabstractindextext{%
\begin{abstract}
Traffic speed is a key indicator for the efficiency of an urban transportation system. Accurate modeling of the spatiotemporally varying traffic speed thus plays a crucial role in urban planning and development. This paper addresses the problem of \textit{efficient} fine-grained traffic speed prediction using big traffic data obtained from static sensors. Gaussian processes (GPs) have been previously used to model various traffic phenomena, including flow and speed. However, GPs do not scale with big traffic data due to their cubic time complexity. In this work, we address their efficiency issues by proposing \textit{local} GPs to learn from and make predictions for correlated subsets of data. The main idea is to quickly group speed variables in both spatial and temporal dimensions into a finite number of clusters, so that future and unobserved traffic speed queries can be heuristically mapped to one of such clusters. A local GP corresponding to that cluster can then be trained on the fly to make predictions in real-time. We call this method \textit{localization}. We use non-negative matrix factorization for localization and propose simple heuristics for cluster mapping. We additionally leverage on the expressiveness of GP kernel functions to model road network topology and incorporate side information. {Extensive experiments using real-world traffic data collected in the two U.S. cities of Pittsburgh and Washington, D.C.,} show that our proposed local GPs significantly improve both runtime performances and prediction accuracies compared to {the baseline global and local GPs}.
\end{abstract}

}

\maketitle

\IEEEdisplaynontitleabstractindextext

\IEEEpeerreviewmaketitle

\section{Introduction}
\label{sect:intro}

Big data captured in densely populated urban environments can provide multi-scaled perspectives at the complex behaviors of urban systems in both space and time. Recent advances in big data technologies such as sensor networks and the Internet of Things (IoT) have accelerated the pace of spatiotemporal data collection in urban settings at ever finer-grained scale. Such wealth of data can be turned into valuable knowledge and insights that can be used to make cities more efficient, safer and enhance the living standard of urban residents. This is a significant utility of big data for social good as it has been forecast that, by 2050, 66\% of the world's population will be urban dwellers \cite{heilig2012world}.

Traffic speed is a key measure of the efficiency of a city's transportation system and the mobility of the urban residents. Accurate modeling and prediction of traffic speed in a city are therefore crucial to the city's intelligent transportation systems (ITS) \cite{zhang2008forecasting, xie2010gaussian}. Traffic speed data are typically obtained from two main sources: one from GPS trajectories generated by moving vehicles equipped with GPS trackers (e.g., taxicabs), and another from static traffic readers or sensors located at fixed locations (e.g., traffic cameras or loop detectors). GPS trajectories are often used as active mobile probes that can directly measure travel times and speeds along road segments \cite{ide2009travel, castro2012urban, ferreira2013visual, wang2014travel, poco2015exploring}. However, using such active probes also incurs high measurement variance due to inconsistent driving behaviors and lack of control over route choices. Hence, a critical mass of probes is needed for each road segment {to obtain} reliable measurements. Meanwhile, static traffic sensors typically provide sparse spatial coverage due to their high installation and maintenance costs. This leaves many road segments uncovered and unobserved and makes it hard to accurately infer traffic speed. Indeed, {recent surveys have} indicated that in most modern cities, only a few main roads have loop detectors installed \cite{schafer2002traffic, chen2014understanding}. This paper examines the latter source of traffic data (i.e., static sensors) for {fine-grained} traffic speed prediction, where ``fine-grained'' here means extensive spatial coverage and fine temporal scales.



In this paper, we address the problem of fine-grained {traffic speed} modeling and prediction in \textit{real-time}. With fast and reliable traffic prediction, travelers can optimize their routes dynamically. Traffic management personnel can also use such information to quickly develop proactive traffic control strategies and make better use of the available transportation resources. Although many navigation systems currently provide live traffic information for routing services, their coverage is limited to major road segments and lacks the predictive capabilities of \textit{future} traffic conditions based on recent observations and historical data \cite{chen2014understanding, poco2015exploring}. In addition, traffic speed in densely populated urban areas is often subject to short-term random fluctuations and perturbations due to exogenous events such as weather conditions, emergencies or traffic incidents \cite{chengaussian}. As a result, we focus on \textit{short-term} traffic prediction in this work\footnote{``Short-term'' can be subjectively defined based on the temporal scale of the sensor readings.} because we find the problem more realistic and challenging.

Gaussian processes (GPs) have been repeatedly demonstrated to be an effective tool for modeling and predicting various traffic phenomena such as mobility demand \cite{chengaussian}, traffic congestion \cite{liu2013adaptive}, short-term traffic volume \cite{xie2010gaussian}, travel time \cite{ide2009travel}, and pedestrian and public transit flows in urban areas \cite{neumann2009stacked}. Indeed, comparative studies on short-term traffic volume prediction showed that GPs outperform other methods such as autoregressive integrated moving average, support vector machine, and multilayer feedforward neural network for the task \cite{xie2010gaussian, zhang2008forecasting}. A particularly attractive feature of GPs is their fully non-parametric Bayesian formulation, which allows for explicit probabilistic interpretation of the model outputs and confidence interval estimations \cite{snelson2007local, xie2010gaussian, chengaussian}. Unfortunately, GPs admit cubic time complexity in the size of the training data. This has been a major limiting factor for the adoption of GPs to model and infer big traffic data, particularly for real-time applications \cite{liu2013adaptive, luttinen2012efficient, chengaussian, zhang2013rass}.

We address the problem of efficient GPs for real-time traffic speed prediction based on the idea of clustering spatiotemporal traffic data into ``local'' subsets of correlated traffic patterns. We call such clustering \textit{localization} \cite{snelson2007local, nguyen2009local, zhang2014fine}. From each subset, a local GP can be trained to make predictions of future traffic queries that could be heuristically mapped to it using some similarity measure. Speed in each local subset is assumed to have similar behaviors through space and time. To this end, we propose to use non-negative matrix factorization (NMF) for fast localization. The idea of using local GPs to infer data of clustered nature is not entirely new. Indeed, Snelson and Ghahramani \cite{snelson2007local} first proposed local GPs for non-linear regression tasks in the biological domain, where clustering is done based on similarity of the responses in the training data. In this work, our adoption of the idea using NMF for efficient traffic speed prediction is novel to the best of our knowledge. We are able to empirically show significant improvements in both runtime performances and prediction accuracies in diverse urban and geospatial settings using the proposed approach compared with baseline methods. Thus, this work can be considered as a hybridization of \cite{xie2010gaussian} that uses GPs for short-term traffic flow prediction and \cite{snelson2007local} that uses the idea of clustering similarly behaved data to train local GPs in order to improve their efficiencies.

In addition, we model traffic speed as spatiotemoporal GPs on road networks, by taking advantage of the expressiveness of the GP kernel functions. Such expressiveness allows us to model the topology and directedness of the road network, as demonstrated by Yu and Chu \cite{yu2008gaussian} for generic networked data. We further take advantage of the \textit{additive kernel} feature of GPs \cite{duvenaud2011additive} to incorporate \textit{side information} into the model, where side information can be any spatial feature of the road network that affects traffic speed through it. Through empirical experiments, we show that there exists an intrinsic tradeoff between model expressiveness and computational efficiency. Model expressiveness translates into more accurate predictions at the cost of increased runtime. In practice, one needs to consider carefully such tradeoff and chooses the most relevant side information to the traffic phenomenon being modeled.

We summarize our main contributions as follows:
\begin{itemize}
\item We develop local Gaussian processes for efficient traffic speed prediction in real-time by using non-negative matrix factorization for clustering of speed in both space and time (i.e., localization).
\item We take advantage of the expressiveness of Gaussian process kernel functions to model traffic speed through directed road networks and incorporate side information features via additive kernel.
\item We perform comprehensive experiments to evaluate our approach using real-world traffic data and demonstrate significant improvements in both runtime and prediction accuracies of using the proposed local GPs against the baseline methods. 
\end{itemize}

The rest of the paper is organized as follows. {In Section \ref{sect:related-work}, we first review recent related works. Section \ref{sect:statement} presents our problem statement, followed by an overview of our solution methodology in Section \ref{sect:overview}. We  describe the NMF and spatiotemporal GPs components of our methodology in Sections \ref{sect:mf} and \ref{sect:gp}, respectively. We then present our experiments in Section \ref{sect:experiments}. Finally, we conclude in Section \ref{sect:conclusion}.}

\section{Background and Related Work}
\label{sect:related-work}

\textbf{Traffic speed data.} Speed modeling is a diverse research area due to a large variety of available metrics and measurement tools (e.g., traffic cameras, GPS traces, speed sensors, etc.) as well as modeling goals. Our work is most closely related to the area of congestion and flow estimation. Congestion and traffic speed estimation has been studied using various mathematical tools, ranging from flow patterns \cite{li2007traffic} to Markov chain forecasting \cite{sirvio2008spatio}, path oracles for spatial networks \cite{sankaranarayanan2009path}, and shortest path and distance queries on road networks \cite{ide2009travel, zhu2013shortest, wang2014travel}. Among those, there are generally two main categories of traffic flow data: (1) \textit{dynamic} traffic measurements obtained from GPS trajectories or low-bandwidth cellular updates associated with individual vehicles \cite{ide2009travel, castro2012urban, wang2014travel, poco2015exploring}, and (2) \textit{static} traffic sensor readings associated with fixed locations (e.g., traffic cameras or sensor networks) \cite{bacon2011using, bottero2013wireless, kafi2013study}. In this respect, our work models data of the second category.

\textbf{Predictive modeling of traffic speed.} Spatiotemporal correlation structure of traffic data can be exploited to predict the speed over unobserved road segments at any time using the observed data at the sensors' locations. Existing Bayesian filtering frameworks \cite{wang2005real, chen2011real} that utilize various handcrafted parametric models to predict traffic flows along highway stretches can only correlate with adjacent highway segments. Thus, their predictive performances could be compromised when the actual spatial correlation spans multiple segments. Moreover, their strong Markov assumption makes these models ungeneralizable to arbitrary road network topology with complex correlation structure. Existing multivariate parametric models \cite{kamarianakis2003forecasting, min2011real} do not quantify uncertainty estimates of the predictions and impose rigid and unrealistic spatial locality assumptions.

\textbf{Gaussian processes for traffic speed.} We model traffic speed as a spatiotemporal Gaussian process (GP) that characterizes the spatiotemporal correlation structure of the phenomenon over a defined road network structure. A major computational advantage of GP is its fully non-parametric Bayesian formulation. This allows for explicit probabilistic interpretation of the model outputs and estimation of predictive uncertainty \cite{rasmussen2004gaussian}. Neumann \textit{et al.} \cite{neumann2009stacked} maintained a mixture of two independent GPs for traffic speed prediction, such that the correlation structure of one GP utilizes road segment features and that of the other GP depends on manually specified relations. Xie \textit{et al.} \cite{xie2010gaussian} {used} GPs to predict the time series of traffic volume over four U.S. highways, and asserted GPs' superior performance over other parametric alternatives. Liu \textit{et al.} \cite{liu2013adaptive} used GPs to model uncertain congestion environments for adaptive vehicle routing. More recently, Chen \textit{et al.} \cite{chengaussian} applied GPs for urban mobility demand sensing in a decentralized and distributed fashion. All these GPs (except for \cite{chengaussian}) do not scale with big traffic data for real-time applications because of their high levels of complexity. In contrast to the distributed GPs proposed in \cite{chengaussian}, our approach is simpler and does not rely on complex decentralized mechanism. 

\textbf{Spatiotemporal clustering.} Clustering techniques have been used to analyze various traffic phenomena. For example, Weijermars \cite{Weijermars2007} applied a hierarchical clustering algorithm to identify typical urban traffic patterns that serve as basis for traffic forecasting. Jiang \textit{et al.} \cite{jiang2012clustering} proposed a framework to cluster the spatiotemporal mobility patterns in urban areas by combining principal component analysis and $K$-means clustering. A common theme is that they employ hard clustering methods that assume each data point can only belong to a cluster. In our work, we relax this assumption by employing NMF \cite{Lee1999,Cichocki2009}, which assumes soft memberships to clusters. Indeed, Ding \emph{et al.} \cite{Ding2005} have shown that, by imposing certain constraints, NMF translates to ``soft'' $K$-means or spectral graph cuts. We also put NMF into a novel application to localize training data for local GPs, making our approach scalable to big data. Our approach also offers a simpler and more generic alternative to the sparsification of GP kernels \cite{snelson2007local,Cao2013}.

\textbf{Local Gaussian processes.} {The idea of localizing training data by clustering in order to learn local GPs has been advocated by several researchers. Snelson and Ghahramani \cite{snelson2007local} developed a local GP approach by dividing the training data into (disjoint) blocks via a simple farthest-point clustering. Nguyen \textit{et al.} \cite{nguyen2009local} proposed a local GP for online regression, where the training data are incrementally partitioned into local regions. For each local region, an individual local GP is trained, and prediction is performed by weighting the nearby local models. While our approach shares similar goals to those, our NMF-based localization is done on the response (i.e., speed) space instead of the feature space. Doing so enables us to build more accurate local GPs, each specializing in a specific traffic response regime.}

{\textbf{Urban computing.} Following the general framework of urban computing research established by Zheng \textit{et al.} \cite{zheng2014urban}, in the urban sensing step, traffic speed data are obtained by fusing public sources of information with real-time speeds crowd-sourced from participating ``floating cars''. In the data management step, GIS shapefiles of the road networks are merged with the collected speed readings to derive features and responses for data analytics and modeling. In the data analytics step, efficient local GPs are used to make real-time inferences of unobserved and future speed values. In the service providing step, the inferred speeds are fed into navigation systems for efficient real-time routing and accurate travel time estimates. Thus, our urban data source is a cross-domain fusion of public sources (i.e., shapefiles for road networks and features, historical traffic flow data from transportation authorities) and privately crowd-sourced speed readings from floating cars. This can be considered as feature-level-based direct concatenation data fusion method according to Zheng \cite{zheng2015methodologies}.}

\begin{figure*}[t]
\centering
\includegraphics[width=0.68\textwidth]{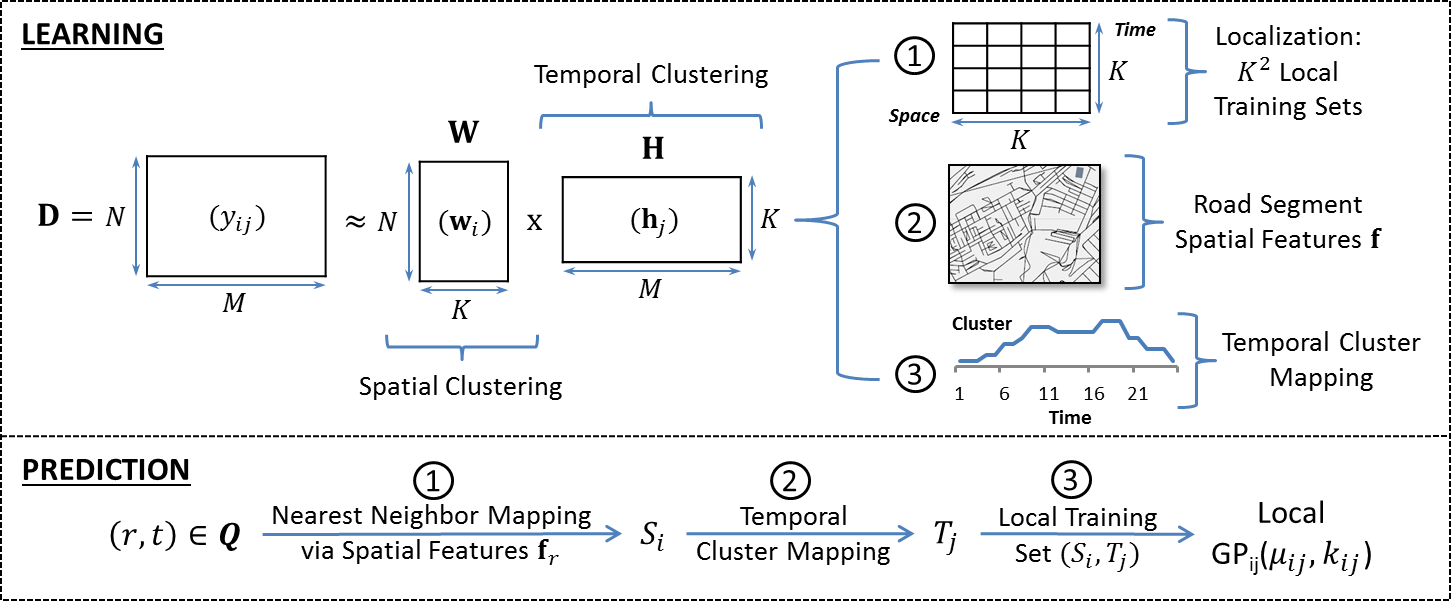}
\caption{The proposed framework for efficient spatiotemporal inference of traffic speed using NMF and local GPs.}
\label{fig:framework}
\end{figure*}

\section{Problem Statement}
\label{sect:statement}
A city's \textit{road network} is a system of interconnected segments and points that represents the land transportation network of a given urban area. A road network can thus be naturally modeled using a graph data structure $G = (V, E)$, where the set of edges $E$ represents the road segments and the set of nodes $V$ represents the intersections (points) among those segments. For many cities around the world, detailed road networks are often made publicly available (typically as GIS shapefiles) by the city's transportation authorities. Moreover, these shapefiles typically contain useful information about the road features such as speed limits, number of lanes, segment length, road type, etc.
	
Suppose we have a road network $G$ and a subset $S \subset E$ of road segments is installed with some form of traffic sensors. Suppose we also have recent observations $\mathbf{D}$ of vehicular travel speeds measured by those sensors at a certain temporal granularity level $\Delta$ (i.e., the sampling interval) along the segments in $S$. Let $r \in E$ be a road segment and $\vec{v}_r$ be the observed speed over $r$, which is inherently a directional quantity (e.g., northbound or southbound). 

\begin{table}[!t]
  \centering
  \caption{Summary of notations used in the paper.}
    \begin{tabular}{ll}
    \hline
      Notation & Description \\
    \hline
    $G, V, E, S$ & Road network $G = (V, E)$, and subset of segments \\
                 & $S \subset E$ that have traffic sensors installed \\
    $\mathcal{S}, \mathcal{T}$ & Set of spatial contexts ($\mathcal{S} \equiv E$) and temporal contexts\\
                 & (e.g., time of the day), respectively \\
    $\mathbf{D}, \mathbf{W}, \mathbf{H}$ & Matrix of observed speeds and its factors,\\
                 & i.e., $\mathbf{D} \approx \mathbf{W} \times \mathbf{H}$ \\
    $N, M, K$ & Dimensions of $\mathbf{D}$ $(N \times M)$, $\mathbf{W}$ $(N \times K)$ and \\
                 & $\mathbf{H}$ $(K \times M)$, where $N = |S|$ and $M = |\mathcal{T}|$ \\
    $\mathbf{Q}$ & Set of traffic speed queries: $\mathbf{Q} = \{(r, t)\}$,\\
                 & where $r \in E$ and $t \in \mathcal{T}$ \\
    $\mathbf{X}$ & Space of spatiotemporal contexts: $\mathbf{X} = \mathcal{S} \times \mathcal{T}$ \\
    $\mathbf{Y}$ & Observed speeds in $\mathbf{D}$, i.e., $\mathbf{D} = (y_{ij})$ \\
    $S_i, T_j$ & Spatial and temporal cluster label $(1 \leq i, j \leq K)$ \\ 
    $k$, $K(\mathbf{X}, \mathbf{X})$ & GP kernel function and covariance matrix\\ 
    $\mathbf{f}_u$, $\mathbf{f}_{(u, v)}$ & Side information: node-wise ($\mathbf{f}_u$) and edge-wise ($\mathbf{f}_{(u, v)}$) \\
    $\Delta$, $W$ & Temporal interval and sliding window\\ 
    \hline
    \end{tabular}
  \label{tab:notations}%
\end{table}%

Given $\mathbf{D}$ and a set $Q \subseteq E$ of querying segments, we seek to answer the following questions:
\begin{enumerate}
\item What are the expected traffic speeds along the segments in ${Q}$ \textit{not} covered by traffic sensors at the current time? We call this the \textit{spatial inference} task.
\item What are the expected traffic speeds along \textit{all} the segments in ${Q}$ in the \textit{near future}\footnote{``Near future'' or ``short-term'' prediction is subjectively defined in this paper as less than $10$ sampling intervals.}? We call this the \textit{temporal prediction} task.
\end{enumerate}

The spatial inference task arises because the spatial coverage of traffic sensors in a city's road network is typically sparse, which may be attributed to their high installation and maintenance costs \cite{schafer2002traffic,chen2014understanding}. The short-term temporal prediction task arises from many real-world applications such as real-time vehicle routing, where new routes are continuously being calculated in light of current and predicted traffic speed information \cite{xie2010gaussian, liu2013adaptive}. Thus, having answers to these questions are the necessary conditions for the solutions to many real-world problems in urban settings, where accurate and fine-grained prediction of the city's spatiotemporally varying traffic speed is crucial.

Table \ref{tab:notations} summarizes the important notations used in the paper as well as their relations.

\section{Solution Overview}
\label{sect:overview}

We address the efficiency issues of using spatiotemporal GPs for learning and predicting large-scale speed data. We draw inspiration from Tobler's first law of geography---``Everything is related to everything else, but \textit{near} things are \textit{more related} than distant things'' \cite{tobler1970computer} to cluster the recently observed traffic speeds in both space and time into ``local'' sets of training data. Each of those subsets corresponds to a local GP. We call such clustering \textit{localization} for short.

Let $\mathbf{Q} = \{(r, t)\}$ be a set of querying road segments at a future time $t$. For each segment $r \in \mathbf{Q}$, we just need to learn a local GP using the segments ``near to'' $r$ \textit{w.r.t. the observed speeds} in order to make a good enough inference of $r$. Likewise, given a future time $t$, we just need to know the data points that are ``related to'' $t$ (w.r.t. the speed) in order to predict those at $t$. We use clustering to quantify such nearness and relatedness in space and time. We propose to use non-negative matrix factorization (NMF) for localization as spatiotemporal clustering is naturally obtained through factorizing the matrix of observed speeds $\mathbf{D}$. The meaning of ``local'' here is the subset of segments and time points in $\mathbf{D}$ that are assumed to have similar speeds to $(r, t)$.

The gain in efficiency comes from the use of a much smaller subset of training data for each local GP, which could be further sped up using parallelization. In addition, using \textit{more relevant} training data could even improve prediction as will be demonstrated. Fig. \ref{fig:framework} illustrates the proposed framework for efficient spatiotemporal inferences for big traffic data using local GPs. The framework consists of two components: \emph{learning} and \emph{prediction}.

\textbf{Learning.} Let $\mathbf{D} = (y_{ij})$ be a matrix of dimension $N \times M$, where $y_{ij}$ is an observed speed value along segment $i$ at time discrete time step $j$, $N = |S|$ is the total number of road segments, and $M = |\mathcal{T}|$ is the total number of regular intervals sampled per day by traffic sensors. The learning process consists of three steps:
\begin{description}
\item[\textbf{Step 1}] We factorize $\mathbf{D}$ into matrices $\mathbf{W} \in \mathbb{R}^{N \times K}_{\geq 0}$ and $\mathbf{H} \in \mathbb{R}^{K \times M}_{\geq 0}$, where $K \ll N, M$. We call $K$ the number of spatial/temporal clusters of $\mathbf{D}$. That is, we could divide the road segments in $S$ into $K$ spatial clusters of similar traffic patterns throughout $\mathcal{T}$ and, likewise, we could divide $\mathcal{T}$ into $K$ temporal clusters of similar traffic patterns throughout $S$. Thus, there are $K^2$ such spatiotemporal clusters, each corresponding to a local training set of a local GP.
\item[\textbf{Step 2}] We normalize $\mathbf{W}$ row-wise. For each row $\mathbf{w}_i$ $(1 \leq i \leq N)$ of $\mathbf{W}$ that corresponds to a road segment $r_i$, we probabilistically assign $r_i$ to one of $K$ spatial clusters using the probability vector $\mathbf{w}_i$. Each $r_i$ also has a vector of spatial features $\mathbf{f}_i$ that is used for spatial clustering mapping.
\item[\textbf{Step 3}] We normalize $\mathbf{H}$ column-wise. For each column $\mathbf{h}_j$ $(1 \leq j \leq M)$ of $\mathbf{H}$ that corresponds to a time step $t_j$, we probabilistically assign $t_j$ to one of $K$ temporal clusters using the probability vector $\mathbf{h}_j$. We call this step \textit{temporal cluster mapping}.
\end{description}

Step $2$ and $3$ perform ``soft assignment'' (i.e., probabilistic mapping) of each road segment and time interval to their respective cluster member. In this respect, NMF is essentially analogous to performing simultaneous clustering  on the rows and columns of $\mathbf{D}$, and probabilistically assigning each row and column vector of $\mathbf{D}$ to their respective cluster member. Because the rows of $\mathbf{D}$ represent the observed traffic patterns over $\mathcal{T}$ at specific road segments, we interpret Step $2$ as \textit{spatial clustering} of road segments according to the similarities of traffic patterns over time. Likewise, each column of $\mathbf{D}$ represents the observed traffic pattern over $S \subset \mathcal{S}$ at certain time interval. Therefore, Step $3$ can be interpreted as \textit{temporal clustering} of time intervals according their similarities of traffic patterns over space. Because the same $K$ is used for both spatial and temporal clustering, we conceptualize such localization as binning the training data $\mathbf{D}$ into $K \times K$ partitions, where each of the partitions (shown as grid cells) is a ``local'' set of training data that have similar traffic pattern in space and time. This concept of localization is illustrated in Step $1$ of Fig. \ref{fig:framework}.

\textbf{Prediction.} Given a query pair $(r, t) \in \mathbf{Q}$, where $t$ is some future time, prediction involves the following steps:
\begin{description}
\item[\textbf{Step 1}] We compare the spatial feature vector $\mathbf{f}_r$ of $r$ with each $\mathbf{f}_s$ of $s$, $\forall s \in {S}$ using the Euclidean distance. We choose the nearest segment $s^* \in {S}$ to $r$. From Step $2$ in Learning, we know which spatial cluster $s^*$ belongs to, here denoted as $S_i$ $(1 \leq i \leq K)$. We deterministically assign $r$ to $S_i$. We call this step \textit{nearest neighbor mapping}.
\item[\textbf{Step 2}] Given $t \in \mathcal{T}$, we simply look up which temporal cluster label $T_j$ ($1 \leq j \leq K$) it belongs to using the temporal cluster mapping (derived in Step 3 of Learning) and deterministically assign $t$ to $T_j$.
\item[\textbf{Step 3}] Given the cluster labels $S_i$ and $T_j$ of $(r, t)$, we retrieve the corresponding local training set $(S_i, T_j)$, train the local GP$(i, j)$ model and make a spatiotemporal inference for $(r, t)$.
\end{description}

For convenience, we shall hereafter use the term ``spatiotemporal inference'' to collectively refer to both the spatial inference (of unobserved segments) and the temporal prediction (of future traffic speed). Each local GP$(i, j)$ can be further extended to consider the network structure and topology in its spatial ``locality'', as well as incorporate side information of the road segments via the its kernel function (see Section \ref{sect:gp}). We shall also use the term ``\textit{global} GP'' to refer to the GP model whose training set is sampled uniformly at random from $\mathbf{D}$ without localization. 

\section{Non-negative Matrix Factorization for Localization}
\label{sect:mf}

\subsection{Preliminaries}

Non-negative matrix factorization (NMF) is a popular technique for decomposing data into latent (hidden) components with physical meaning and interpretations \cite{Lee1999,Cichocki2009}. It has been widely used in dimensionality reduction, object detection, latent clustering, and blind source separation, involving image, text and signal data \cite{Shashua2006,Sun2006,Cichocki2009}. In this work, we use NMF to decompose matrix $\mathbf{D}$ into two non-negative matrices $\mathbf{W}$ and $\mathbf{H}$ that represent the spatial and temporal clusters of speed values in $\mathbf{D}$, respectively. These two matrices are then used for 
the localization of GPs during the training and prediction phases.

More formally, NMF seeks to approximate $\mathbf{D} \in \mathbb{R}^{N \times M}_{\geq 0}$ by a product of $\mathbf{W} \in \mathbb{R}^{N \times K}_{\geq 0}$ and $\mathbf{H} \in \mathbb{R}^{K \times M}_{\geq 0}$ (i.e., $\mathbf{D} \approx \mathbf{W} \times \mathbf{H}$), where $K$ is the number of clusters. Note that usually $K \ll \min(N, M)$. The non-negativity constraint imposed on the two matrices serves to provide meaningful interpretations for the spatial and temporal clusters. That is, each row of $\mathbf{W}$ can be interpreted as the \emph{degrees of membership} to $K$ different spatial clusters. Likewise, each column of $\mathbf{H}$ represents the degrees of membership to $K$ different temporal clusters. 

\subsection{Optimization Objective}
\label{sect:nmf_objective}

The quality of approximating $\mathbf{D}$ by $\mathbf{W} \times \mathbf{H}$ can be measured through various distance functions. In this work, we use the Frobenius norm, which leads to the optimization problem of \emph{minimizing} the loss function $\mathcal{L}$:
\begin{align}
\label{eqn:nmf_loss}
\mathcal{L} &= \frac{1}{2} ||\mathbf{D} - \mathbf{W} \mathbf{H}||_F^2 = \frac{1}{2} \sum_{i,j} \left[ \mathbf{D}_{i,j} - \sum_{k} \mathbf{W}_{i,k} \mathbf{H}_{k,j} \right]^2
\end{align}
where $i \in \{1,\ldots,N\}$, $j \in \{1,\ldots,M\}$, and $k \in \{1,\ldots,K\}$.

To arrive at meaningful spatial and temporal clusters, we further impose \emph{sparsity} constraints to $\mathbf{W}$ and $\mathbf{H}$ via L1-norm penalty. This yields the following regularized loss:
\begin{align}
\label{eqn:sparse_nmf_loss}
\mathcal{L} &= \frac{1}{2} ||\mathbf{D} - \mathbf{W} \mathbf{H}||_F^2 + \lambda \underbrace{\left( \sum_{i,k} \mathbf{W}_{i,k} + \sum_{j,k} \mathbf{H}_{k,j} \right)}_{\text{L1-norm penalty}},
\end{align}
where $\lambda > 0$ is the regularization parameter (set to $\lambda=100$). Enforcing sparse $\mathbf{W}$ and $\mathbf{H}$ leads to sparse membership to different clusters, thus improving the model interpretability while retaining approximation quality.

It is also worth noting that $\mathcal{L}$ is convex with respect to the individual matrix $\mathbf{W}$ or $\mathbf{H}$, but not both. As a result, one can only expect to find a stationary point of $\mathcal{L}$, which is not necessarily a globally optimal solution. In the following section, we describe a fast coordinate descent algorithm to find a stationary solution to the optimization problem (\ref{eqn:sparse_nmf_loss}).

\subsection{Coordinate Descent Learning}
\label{sect:nmf_cd}

The key idea of the coordinate descent (CD) method is to update one variable at a time, while keeping the others fixed. The efficiency of the CD procedure has been demonstrated in several state-of-the-art machine learning methods \cite{Fan2008,Friedman2010}. For NMF, the conventional ways of learning $\mathbf{W}$ and $\mathbf{H}$ are largely based on the alternative non-negative least squares (ANLS) framework \cite{Paatero1994}, which converges to stationary points provided each sub-problem can be solved exactly. However, the ANLS-based methods usually take a significant amount of time to find an exact solution for each sub-problem. In contrast, the CD method can efficiently compute reasonably good solution for each sub-problem and move on to the next round \cite{Friedman2010}.

Without loss of generality, we shall focus on the coordinate descent update for entries in $\mathbf{W}$; the update for entries in $\mathbf{H}$ can be similarly derived, i.e., by replacing $\mathbf{D}$ with $\mathbf{D}^\top$ and swapping $\mathbf{W}$ with $\mathbf{H}^\top$. The CD method solves each sub-problem by the following one-variable Newton update:
\begin{align}
\label{eqn:cd_update}
\mathbf{W}_{i,k} \leftarrow \max \left(0, \mathbf{W}_{i,k} - \frac{(\bigtriangledown_{\mathbf{W}} \mathcal{L})_{i,k}}{(\bigtriangledown_{\mathbf{W}}^2 \mathcal{L})_{i,k}} \right),
\end{align} 
where $\bigtriangledown$ and $\bigtriangledown^2$ denote the gradient (i.e., first derivative) and curvature (i.e., second derivative), respectively. The truncation $\max(0, x)$ serves to ensure non-negative $\mathbf{W}$. 

With respect to the regularized loss (\ref{eqn:sparse_nmf_loss}), it is easy to show that the gradient $\bigtriangledown_{\mathbf{W}} \mathcal{L}$ resolves to:
\begin{align}
\bigtriangledown_{\mathbf{W}} \mathcal{L} &= \mathbf{W} \mathbf{H} \mathbf{H}^\top - \mathbf{D} \mathbf{H}^\top + \lambda,
\end{align} 
and in turn the curvature $\bigtriangledown_{\mathbf{W}}^2 \mathcal{L}$ is:
\begin{align}
\bigtriangledown_{\mathbf{W}}^2 \mathcal{L} &= \mathbf{H} \mathbf{H}^\top.
\end{align}

Consequently, the CD update in (\ref{eqn:cd_update}) can be written as:
\begin{align}
\label{eqn:cd_update_full}
\mathbf{W}_{i,k} \leftarrow \max \left(0, \mathbf{W}_{i,k} - \frac{(\mathbf{W} \mathbf{H} \mathbf{H}^\top - \mathbf{D} \mathbf{H}^\top)_{i,k} + \lambda}{(\mathbf{H} \mathbf{H}^\top)_{i,k}} \right).
\end{align} 

It can be seen from (\ref{eqn:cd_update_full}) that the regularization parameter $\lambda$ plays a role in shifting the new $\mathbf{W}_{i,k}$ to a smaller (possibly negative) value. As such, a larger $\lambda$ would foster more (zero) truncation and therefore result in a sparser solution. 

Using the update rule (\ref{eqn:cd_update_full}), we carry out a \emph{cyclic} coordinate descent. That is, we first update all entries in $\mathbf{W}$ in cyclic order, and then update entries in $\mathbf{H}$, and so on. With respect to $\mathbf{W}$, we traverse every cluster $k$, in which we update each variable $\mathbf{W}_{i,k}$ using (\ref{eqn:cd_update_full}). The same applies to each $\mathbf{H}_{k,j}$, with $\mathbf{W}$ swapped with $\mathbf{H}^\top$. The procedure is repeated until a maximum number of iterations (set to $200$) is reached. 

\subsection{Efficiency Considerations}
\label{sect:nmf_efficiency}

The aforementioned CD procedure can be carried out efficiently if certain quantities are pre-computed. Specifically, we calculate and store the matrix products $\mathbf{D} \mathbf{H}^\top$ and $\mathbf{H} \mathbf{H}^\top$ prior to entering the one-variable update loop for $\mathbf{W}$. (Similarly, we pre-compute $\mathbf{D}^\top \mathbf{W}$ and $\mathbf{W}^\top \mathbf{W}$ before updating $\mathbf{H}$). These would incur an additional memory with an order of $\mathcal{O} \left( (N + M) \times K \right)$ and $\mathcal{O} \left( K^2 \right)$, respectively. As such, the total memory complexity of the CD procedure is $\mathcal{O} \left( (N + M + K) \times K \right)$. This, however, is still much smaller than the dimensionality of $\mathbf{D}$ (i.e., $N \times M$). 

Meanwhile, thanks to caching, the time complexity of the CD procedure is \emph{linear} with respect to $N$ and $M$. In particular, the time needed to update all entries in $\mathbf{W}$ within a CD iteration is $\mathcal{O} \left( N \times K^2 \right)$. Similarly, the time for updating $\mathbf{H}$ is $\mathcal{O} \left( M \times K^2 \right)$. Thus, the overall time complexity is thus $\mathcal{O} \left( (N + M) \times K^2 \times T_{max} \right)$ (where $T_{max}$ is the maximum number of iterations). As $K$ and $T_{max}$ are typically small, fixed values that are independent of the problem size, we conclude that the CD procedure is efficient. We empirically demonstrate its efficiency in Section \ref{sect:clustering}.

\subsection{Determining  $K$}
\label{sect:determine_K}

{One practical problem in applying NMF is to determine the optimal number of clusters $K$. In this work, we use {$10$-fold cross validation} (CV) procedure to determine $K$. Specifically, we randomly split all entries $y_{ij}$ of $\mathbf{D}$ into $10$ mutually exclusive folds, and for each CV iteration $f$, we use fold $f$ as validation set for NMF, and the remaining (nine) folds as training set. We then determine the optimal number of clusters by choosing $K$ that gives the highest \emph{fraction of explained variance} score \cite{draper1998regression} averaged over 10 validation sets.}

{For a target (speed) variable $y$ and predicted (speed) variable $\hat{y}$, the fraction of expected variance $R^2(y, \hat{y})$ is:
\begin{align}
\label{eqn:explained_var}
R^2(y, \hat{y}) = 1 - \frac{Var[y - \hat{y}]}{Var[y]},
\end{align}
where $Var[y] = \mathbf{E}[y^2] - (\mathbf{E}[y])^2$ is the variance of $y$. 
} 

{Notably, the fraction of explained variance is a popular metric commonly used to evaluate a regression model \cite{draper1998regression}. For an optimal regression model $\hat{y}$ that perfectly matches the target variable $y$, the variance $Var[y - \hat{y}]$ will be zero, which in turn implies $R^2(y, \hat{y}) = 1$. On the other hand, the most na\"{i}ve regression model is a constant function, which gives $Var[y - \hat{y}] = Var[y]$ and thus $R^2(y, \hat{y}) = 0$. In this case, the prediction $\hat{y}$ tells us nothing about the target $y$, in the sense that $\hat{y}$ does not covary with $y$.}

\section{Spatiotemporal Gaussian Processes for Traffic Speed Modeling}
\label{sect:gp}
\subsection{Preliminaries}
\label{sect:gp-prelim}

Let $\mathcal{S}$ denote the space of spatial contexts (i.e., $\mathcal{S} \equiv E$ in this paper) and $\mathcal{T}$ denote the space of temporal contexts (e.g., information about time of the day). We model the speed over road segment $r \in E$ under varying $t \in \mathcal{T}$ via the function $f : \mathcal{S} \times \mathcal{T} \mapsto \mathbb{R}_{\geq 0}$ that outputs a non-negative speed value for a given $(r, t)$ pair.

We define a spacetime process as a stochastic process indexed by road segments ${r} \in \mathcal{S}$ and temporal labels $t \in \mathcal{T}$:
\begin{equation}
\label{eqn:spacetime-process}
	\{f({r}, t): {r} \in \mathcal{S}, t \in \mathcal{T}\}.
\end{equation}
Thus, for a fixed spacetime location $(r, t)$, $f(r, t)$ is a random variable. It is a fundamental nature of spatiotemporal data that observations at \textit{nearby} locations in space and time are similar \cite{rasmussen2004gaussian}. We need a mathematical model to quantify the extent to which things are related over space and time. Kernel functions provide such an elegant model. For example, given two spacetime locations $(r, t)$ and $(r', t')$, the radial basis function (RBF) kernel has the following form:
\begin{equation}
\label{eqn:rbf-kernel}
	k((r, t), (r', t')) = e^{-{\lVert(r, t) - (r', t')\rVert}/{l^2}}.
\end{equation}

A spatiotemporal Gaussian process (GP) is a stochastic process over an index set $\mathbf{X} = \mathcal{S} \times \mathcal{T}$. It is entirely defined by a mean function $\mu: \mathbf{X} \mapsto \mathbb{R}_{\geq 0}$ and a covariance ({kernel}) function $k: \mathbf{X} \times \mathbf{X} \mapsto \mathbb{R}$. These two functions are chosen such that they jointly define a multivariate normal distribution whenever we draw $f | \mathbf{X}$ from a GP($\mu, k$) on a finite set of spacetime locations $\mathbf{X} = \{x_1, \ldots, x_T\}$:
\begin{equation}
	f | \mathbf{X} \sim \mathcal{N}(\mu(\mathbf{X}), K(\mathbf{X}, \mathbf{X})),
\end{equation}
where $\mu(\mathbf{X})_i = \mu(x_i)$ and $[K(\mathbf{X}, \mathbf{X})]_{ij} = k(x_i, x_j)$.

By this construction, $\mu(\mathbf{X})$ is a $T$-dimensional non-negative vector and $K(\mathbf{X}, \mathbf{X}) \in \mathbb{R}^{T \times T}$ is a positive semidefinite covariance matrix. We now assume that $f$ is sampled probabilistically from a GP prior $f \sim P(f)$  \cite{rasmussen2004gaussian}.  A GP prior is fully specified by its mean function:
	$$\mu(r, t) = \mathbb{E}[f(r, t)],$$
its covariance (or {kernel}) function:
\begin{equation*}
\begin{aligned}
	k((r, t),(r', t')) &= \mathbb{E}[(f(r, t) - \mu(r, t))(f(r', t') - \mu(r', t'))] \\
	&= \mathrm{Cov}((r, t),(r', t')),
\end{aligned}
\end{equation*}
and observation noise with variance $\sigma^2$.

A major computational benefit of GPs is that the posterior can be computed in a closed form.  Suppose we have collected recent speed observations $\mathbf{Y} = [y_1,\ldots,y_T]^\top$ at $\mathbf{X} = [(r_1, t_1), \ldots,(r_T, t_T)]$. We can write the posterior distribution of $f$ given $\mathbf{X}$ and $\mathbf{Y}$ also as a GP with mean:
\begin{small}
\begin{eqnarray}
	\mu_{\mathbf{Y}, \mathbf{X}}(r, t) = \mu(r, t) + \hat{k}_{\mathbf{X}}(r, t)^\top (\hat{K}_{\mathbf{Y}, \mathbf{X}} + \sigma^2 I )^{-1} (\delta \mathbf{Y})^\top
\label{eqn:mu}
\end{eqnarray}
\end{small}
and covariance $k_{\mathbf{Y}, \mathbf{X}}((r, t),(r', t'))=$
\begin{small}
\begin{eqnarray}
	k((r, t),(r', t')) - \hat{k}_{\mathbf{X}}(r, t)^\top
(\hat{K}_{\mathbf{X}} + \sigma^2 I )^{-1} \hat{k}_{\mathbf{X}}(r', t'),
\label{eqn:cov}
\end{eqnarray}
\end{small}
where $\delta \mathbf{Y}$  is the deviation of $\mathbf{Y}$ from its prior mean:
$$\delta \mathbf{Y} = [y_1-\mu(r_1, t_1),\ldots,y_T-\mu(r_T, t_T)]^\top,$$  
$\hat{k}_{\mathbf{X}}(r, t)$ is a column vector of the kernel values between $(r, t)$ and each observed location in  $\mathbf{X}$:
$$\hat{k}_{\mathbf{X}}(r, t) = [k((r_1, t_1), (r, t)),\ldots,k((r_T, t_T), (r, t))]^\top \in \mathbb{R}^{T},$$
and $\hat{K}_{\mathbf{X}}$ is the Gram matrix of all locations  in $\mathbf{X}$:
$$\hat{K}_{\mathbf{X}} =  [k((r_i, t_i),(r_j, t_j))]_{i, j \in [1, \ldots, T]} \in \mathbb{R}^{T \times T}.$$
The posterior variance of $f(r, t)$ is $k_{\mathbf{Y}, \mathbf{X}}((r, t),(r, t))$.



Inference of continuous values with GP prior is known as GP regression (or \textit{kriging}). When concerned with a general GP regression, it is assumed that for a GP $f$ observed at location $(r, t)$, $f(r, t) | \Theta$ is just one sample from the multivariate normal distribution of dimension $|\mathbf{X}|$, where $\Theta$ is the set of hyper-parameters of the kernel function $k((r, t),(r', t'))$. Thanks to its non-parametric nature, training a GP reduces to estimating $\Theta$ via the marginal likelihood function. Having identifying $\Theta$, spatiotemporal inference $f(r', t')$ becomes a matter of sampling from the posterior distribution. A major computational bottleneck of GP is its $\mathcal{O}(|\mathbf{X}|^3)$ time complexity, which makes it impractical for large-scale spatiotemporal data \cite{rasmussen2004gaussian, luttinen2012efficient, chengaussian}.

\subsection{Kernel Functions for Road Networks}
\label{sect:gp-networks}

Let $G = (V, E)$ be a directed graph representing a road network. $G$ is directed because traffic on a road segment could possibly be one-way. On two-way segments, the corresponding links of $G$ become bidirectional. Let $\mathbf{Y} = \{y_{(u, v)} : (u, v) \in E\}$ be the speed values that we wish to model. An important nature of networks is that $\mathbf{Y}$ are highly correlated on known node and edge features. Following Yu and Chu \cite{yu2008gaussian}, let $f: V \times V \mapsto \mathbb{R}_{\geq 0}$ be a GP($\mu, k$), then the kernel function between $(u, v)$ and $(u', v')$ can be written as:
\begin{equation}
\label{eqn:link-kernel}
	k((u, v), (u', v')) = k(u, u')k(v, v'),
\end{equation}
where $k: V \times V \mapsto \mathbb{R}$ is some kernel function between the nodes. Since a random function $f$ drawn from GP$(\mu, k)$ is generally asymmetric, i.e., $f(u, v) \neq f(v, u)$, traffic directions along the links in $G$ are automatically modeled.

Let $u, v \in V$ be identified by their respective pair of longitude and latitude coordinates $(u_x, u_y)$ and $(v_x, v_y$), then equation \eqref{eqn:link-kernel} becomes:
\begin{equation}
\begin{aligned}
\label{eqn:coord-kernel}
	& k((u, v), (u', v')) \\
	&= k((u_x, u_y), (u'_x, u'_y)) k((v_x, v_y), (v'_x, v'_y)).
\end{aligned}
\end{equation}

For spatiotemporal data, a natural way to formulate a spacetime kernel is to multiply the spatial kernel $k_s$ {and} the temporal kernel $k_t$ together. This feature is referred to as \textit{separable kernel} of GPs\cite{rasmussen2004gaussian, luttinen2012efficient}. Let $r = (u, v), r' = (u', v') \in E$ and $t$ be a time label, from \eqref{eqn:coord-kernel}, we have:
\begin{equation}
\begin{aligned}
\label{eqn:separable}
	& k((r, t), (r', t')) \\
	&= k_s((u_x, u_y), (u'_x, u'_y)) k_s((v_x, v_y), (v'_x, v'_y)) k_t(t, t').
\end{aligned}
\end{equation}

\subsection{Incorporating Side Information}

We define \textit{side information} as any spatial features of the nodes and edges of $G$ other than the longitude and latitude coordinates of the nodes of $G$, which precisely specify the geolocation of a given edge $(u, v)$ and quantify its geospatial nearness to another edge $(u', v')$. Therefore, side information could be any \textit{other} spatial features of the nodes and edges of $G$ that can be derived from the given GIS shapefile of the road network. We then classify side information into two types: node-wise and edge-wise side information, where \textit{node-wise} side information contains the spatial features of the nodes of $G$ and \textit{edge-wise} side information contains the spatial features of the edges of $G$.

For each road segment $r = (u, v)$, let $\mathbf{f}_u$ and $\mathbf{f}_v$ denote the vectors of node-wise side information of $r$, which are necessarily of the same length. Likewise, let $\mathbf{f}_{(u, v)}$ denote the vector of edge-wise side information of $r$. The set of \textit{all} side information of $r$ is denoted as $\mathbf{f}_r = (\mathbf{f}_u; \mathbf{f}_v; \mathbf{f}_{(u, v)})$. We take advantage of the \textit{additive kernel} feature of GPs \cite{duvenaud2011additive} to incorporate side information into the kernel function. Following \eqref{eqn:separable}, the kernel function between $(r, t)$ and $(r', t')$ knowing their side information $\mathbf{f}_r$ and $\mathbf{f}_{r'}$ is given by:

\begin{equation}
\begin{aligned}
	& k((r, t, \mathbf{f}_r), (r', t', \mathbf{f}_{r'})) \\
	&= k((r, t), (r', t')) + \sum_i k(\mathbf{f}_u^{(i)}, \mathbf{f}_{u'}^{(i)}) k(\mathbf{f}_v^{(i)}, \mathbf{f}_{v'}^{(i)}) \\
	&+ \sum_j k(\mathbf{f}_{(u, v)}^{(j)}, \mathbf{f}_{(u', v')}^{(j)}),
\end{aligned}
\end{equation}
where $i$ and $j$ are the indices of the set of node-wise and edge-wise side information, respectively.

\subsection{Complexity of Local GPs}
\label{sect:gp-computation}

For each local GP$(i, j)$, without incorporating side information, the time complexity is $\mathcal{O}(|\mathbf{X}_{ij}|^3) = \mathcal{O}(|\mathcal{S}_i|^3 \times |\mathcal{T}|_j^3)$. The original sizes of ${S} \subset E$ and the space of temporal contexts $\mathcal{T}$ from matrix $\mathbf{D}$ are $N$ and $M$, respectively. Due to clustering, each local training set $({S}_i, {T}_j)$ has $\mathbb{E}[|\mathcal{S}_i|] = N / K$ and $\mathbb{E}[|\mathcal{T}_i|] = M / K$ training data points on expectation. Thus, the expected time complexity of each local GP$(i, j)$ is $\mathcal{O} ((\frac{N M}{K^2})^3)$. If the prediction phase in Fig. \ref{fig:framework} can be done in parallel for each spatiotemporal cluster $({S}_i, {T}_j)$, then $\mathcal{O} ((\frac{N M}{K^2})^3)$ is the expected time complexity to predict an arbitrary set of queries $\mathbf{Q} = \{(r, t)\}$. Otherwise, if it is done serially, then the worst-case time complexity is $K^2 \mathcal{O} ((\frac{N M}{K^2})^3) = \mathcal{O} (K^2 (\frac{N M}{K^2})^3) = \mathcal{O} (\frac{(N M)^3}{K^4})$, which is still a significant improvement over the original $\mathcal{O} ((N M)^3)$ time complexity of global GPs without side information.

For GPs with side information, the total time complexity is added by the complexity of the kernel function of each ``piece'' of side information, each having complexity of $\mathcal{O}(N^3)$ and $\mathcal{O}((\frac{N}{K^2})^3)$ for global and local GPs, respectively. We will empirically demonstrate in the next section the effects of having side information on the ``wall-clock'' runtime performances of both local and global GPs.

\section{Empirical Evaluation}
\label{sect:experiments}

\begin{figure}[!t]
\centering
\includegraphics[width=1.0\columnwidth]{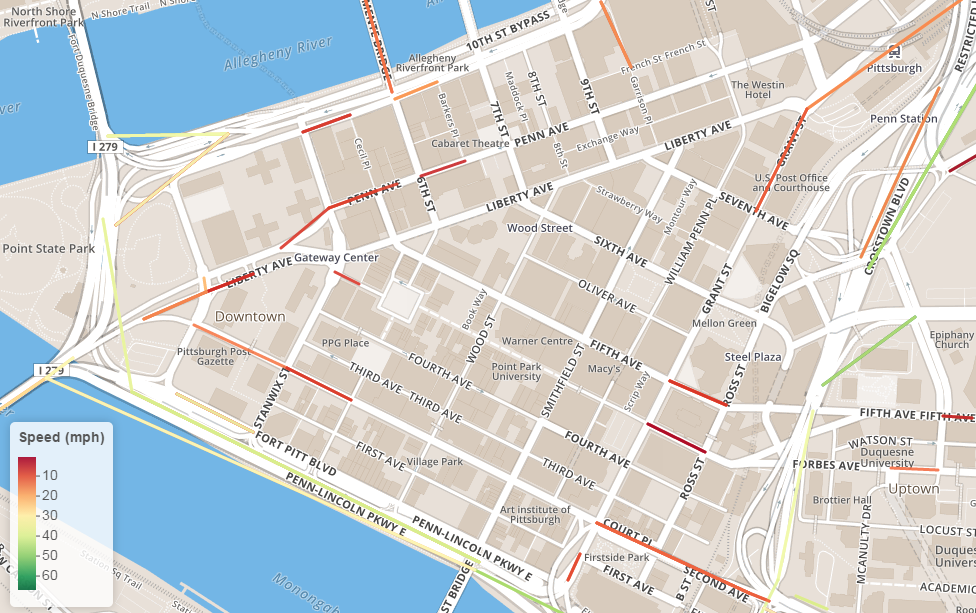}
\caption{Visualization of the speed distribution along road segments covered by our TMC dataset in downtown Pittsburgh on a typical {weekday} in August, 2014 at 8 a.m.}
\label{fig:train-weekday}
\end{figure}

\subsection{Datasets}
\label{sect:dataset}

TMC (Traffic Message Channel) is a technology used to broadcast traffic information in real-time to vehicles through the radio waves. TMC allows for silent delivery of dynamic traffic information, and is often integrated directly into the vehicle's navigation system for real-time estimation of speed and route calculation. {We have acquired, through a commercial vendor of navigation systems, rich TMC datasets that record the average speeds along certain road segments in the two U.S. cities of Pittsburgh, Pennsylvania (P.A.), and Washington, D.C. Our TMC datasets cover a total of $1,190$ and $1,091$ unique road segments in the city's road network of Pittsburgh and Washington, respectively.} Each record is an average speed measurement over a road segment every $5$-minute interval (i.e., $\Delta = 5$ minutes) everyday for the whole summer month\footnote{Traffic pattern typically remains the same during a season\cite{poco2015exploring}, which justifies our choice of data.} of August, 2014. Each speed value also has a direction indicator (e.g., northbound, southbound, eastbound or westbound). Thus, our dataset is a close approximation to the city's traffic sensor network.

{TMC technology fuses real-time traffic information from crowd-sourced networks of ``floating cars'' and mobile devices with public sources of information (e.g., from historical data or transportation authorities). Under normal conditions, when no incidents are reported from crowd-sourced devices, TMC data capture publicly available sources of traffic information. Under irregular conditions, such as traffic incidents or congestion, crowd-sourced information is collected and broadcast to alert drivers in real-time. Still, TMC data can be missing for certain road segments when routing services are not usually called for. This happens typically in the late night or early morning hours. Hence, our data are temporally sparse for each road segment, i.e., there are many missing values in the temporal dimension.}

We downloaded the shapefiles\footnote{The shapefile of Pittsburgh's road network can be downloaded from: \href{http://pittsburghpa.gov/dcp/gis/gis-data-new}{http://pittsburghpa.gov/dcp/gis/gis-data-new}, and Washington's from: \href{http://opendata.dc.gov}{http://opendata.dc.gov}.} representing the two cities' road networks and constructed a connected directed graph $G = (V, E)$ for each. {Our datasets cover approximately $5\%$ and $8\%$ of the city's road network for Pittsburgh and Washington, respectively.} We extract useful spatial features of the road segments in $G$ from the retrieved shapefiles and the network structure of $G$. Table \ref{tab:features} summarizes those spatial features. The table also shows two \textit{network centrality} measures of $G$: (node) degree and (edge) betweenness. Node degree is the (all) degree of a node in the directed network. 
Edge betweenness is the number of shortest paths from all pairs of nodes in the network that pass through a given edge \cite{borgatti2005centrality}. 
Network centralities have been shown to greatly influence on the flow of information and traffic through diverse networked settings \cite{holme2003congestion, borgatti2005centrality, le2013empirical}.

\begin{table*}[!t]
  \centering
  \caption{The extracted spatial features $\mathbf{f}$ of the road segments in Pittsburgh and Washington.}
    \begin{tabular}{rl}
    \hline
      Feature & Description \\
    \hline
    Longitude, latitude & Longitude and latitude coordinates of the two endpoints (nodes) of a segment. \\
    Segment length & Length (in miles) of a segment. \\
    Number of lanes & The number of lanes a segment has in each direction. \\
    Direction & Direction of a segment: northbound, southbound, eastbound, or westbound. \\
    Degree & Degree of two end nodes of an edge (segment). \\
    Betweenness & Edge betweenness centrality of a segment. \\
    One-way & Is this segment one-way? \\
    Road type & One of the $10$ defined types: avenue, boulevard, bridge, lane, place, ramp, road, street, tunnel, and way. \\
    \hline
    \end{tabular}
  \label{tab:features}%
\end{table*}%

Fig. \ref{fig:train-weekday} visualizes the speed distribution over the road segments covered by our TMC data in downtown Pittsburgh on a typical weekday. Speed value along a segment is averaged over observations on all the weekdays in the month at 8 a.m. The figure shows smaller segments in the downtown area tend to have lower speeds during the morning rush hour. Larger segments, on the other hand, are reasonably observed with higher speeds and faster flows.

\begin{figure}[!t]
\begin{center}
	\includegraphics[width=1\columnwidth]{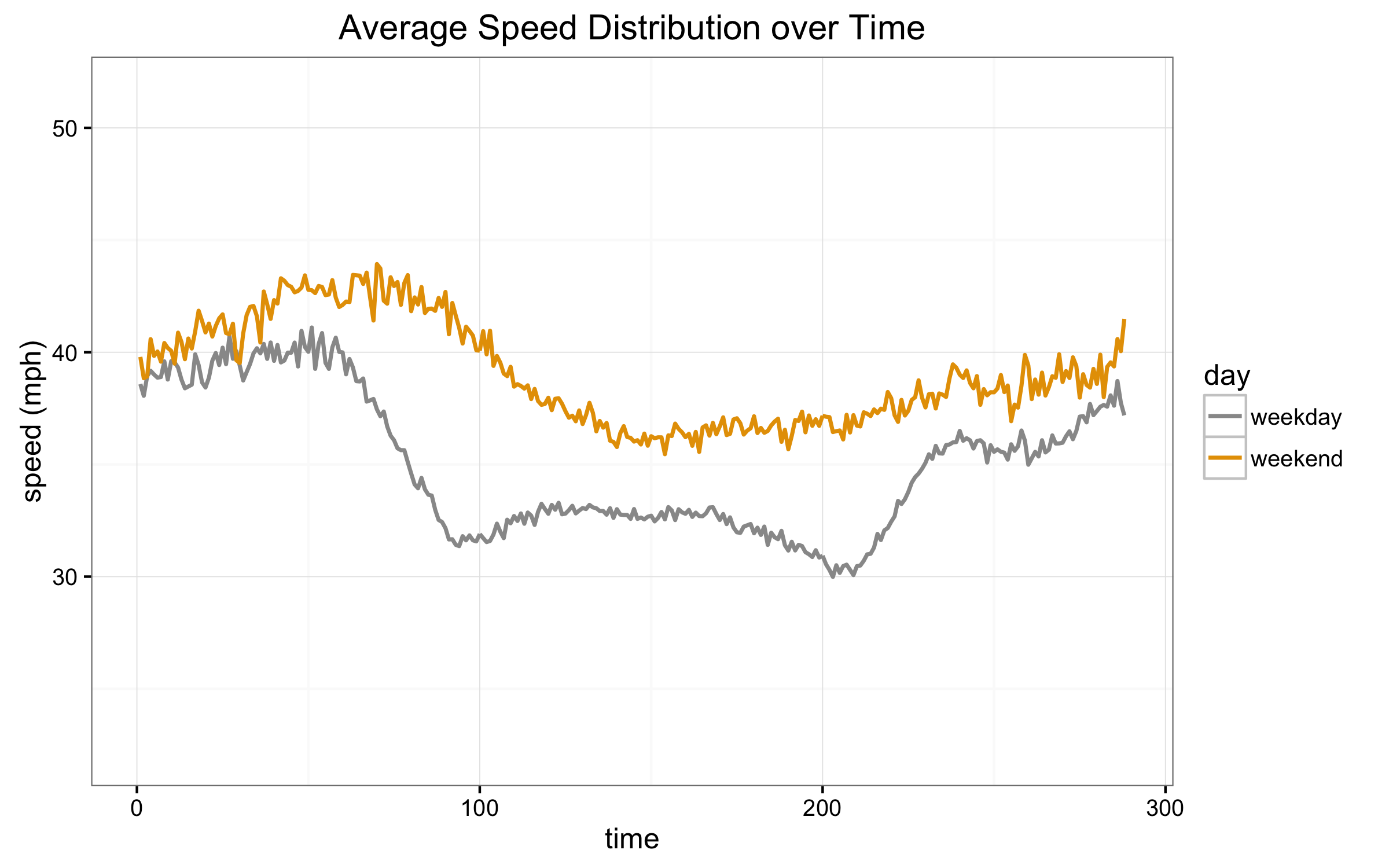}
\end{center}
\caption{Time series of the average speed along road segments in Pittsburgh every $5$-minute interval in August, 2014.}
\label{fig:time-series}
\end{figure}

Fig. \ref{fig:time-series} shows the time series of the average speed on all observed road segments in Pittsburgh during all the weekdays and weekends in the month. The figure clearly shows that traffic speed on the weekend is, on average, faster and less variable than that on the weekday. It also shows the rush hours effects on the weekday: average speed dips around $8$ a.m. (morning rush hour) and $5$ p.m. (evening rush hour) when people commute to work and go home, respectively. The traffic between those two rush hours is generally much slower than in the late evening and early morning. On the weekend, by contrast, traffic is generally slower during the day when people tend to go out. {The data of Washington, D.C., exhibit very similar patterns.}

\subsection{Experiment Design}
\label{sect:design}

Following the observations in Fig. \ref{fig:time-series} and the established procedures in modeling human mobility patterns in urban areas \cite{zhang2008forecasting, xie2010gaussian, jiang2012clustering, ferreira2013visual}, we split the data of each city into two sets: weekday (Monday through Friday) and weekend (Saturday and Sunday). We design the following experiments to measure the performances of our local GP models in diverse spatiotemporal settings using both sets.

For each city, we designate Thursday, August 28, 2014 and Sunday, August 31, 2014 as the test weekday and weekend, respectively. We choose Thursday as a test weekday as previous studies have suggested the inherent differences in urban mobility patterns between Friday and the rest of the weekdays \cite{jiang2012clustering, ferreira2013visual, poco2015exploring}. We call either date the \textit{test day}. For each hour $t \in \{0, 1, \ldots, 22, 23\}$ on each test day (of each city), we designate the \textit{test time} to be $1$--$6$ intervals ahead of $t$, i.e., test time is $t + i \times \Delta$, where $\Delta = 5$ minutes and $1 \leq i \leq 6$. There are $24$ trials per test day, where each trial predicts $6$ test cases. We call each test case an $i$-step ahead prediction and simply denote the test time as $t + i$.

We adopt the ``sliding window'' method proposed in \cite{xie2010gaussian} to collect the training data for each trial $t$, denoted as $\mathbf{D}_t$. Given a test time $t + i$, $\mathbf{D}_t$ is the observations collected from time $t-W$ up to (and including) $t$, where $W$ is the length of the window of observations. For weekday, $W$ is a period of \textit{exactly} $5$ previous weekdays, i.e., $W = 5 \times 24 \times 12 = 1,440$ intervals. For weekend, $W$ is a period of \textit{exactly} $3$ previous weekend days, i.e., $W = 3 \times 24 \times 12 = 864$ intervals. We empirically choose such $W$ for both sets in order to avoid the ``cold start'' problem\footnote{The cold start problem invalidates the factorization of $\mathbf{D}$ if there exists either an entire row of column of $\mathbf{D}$ that admits \textit{all} missing values.} in matrix factorization \cite{Lee1999} due to the temporal sparsity problem of our data. For each trial, $\mathbf{D}_t$ is the observed speeds averaged \textit{over the days} in $W$.

\begin{figure}[!t]
\centering
\includegraphics[width=0.80\columnwidth]{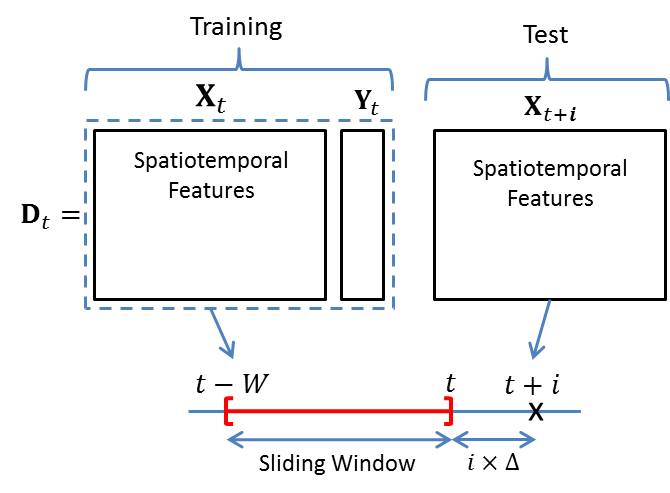}
\caption{The adopted ``sliding window'' experimental design: $t \in \{0, 1, \ldots, 22, 23\}$ on the test day, $t + i$ $(1 \leq i \leq 6)$ denotes the test time. $W$ is the length of the sliding window: $5$ days for weekday and $3$ days for weekend. $\mathbf{D}_t$ denotes the training data containing the features $\mathbf{X}_t$ and the observed speeds $\mathbf{Y}_t$ in $\mathbf{D}_t$ averaged over $24$-hour periods in $W$.}
\label{fig:experiments}
\end{figure}

To evaluate the spatiotemporal inferences of our models, we randomly select $40\%$ of the segments out of the total number of segments as the training set and test on \textit{all} the segments. {Hence, each $\mathbf{D}_t$ is a $(476 \times 288)$- and $(436 \times 288)$-dimensional matrix for Pittsburgh and Washington, respectively.} Fig. \ref{fig:experiments} illustrates our experimental design.

The following models are considered in our experiments:
\begin{enumerate}
\item \textbf{GP} -- global GP without side information;
\item \textbf{GP$^+$} -- global GP with side information;
\item \textbf{LGP} -- NMF-based local GP {without} side information;
\item \textbf{LGP$^+$} -- NMF-based local GP {with} side information;
\item {\textbf{LGR} -- grid-based local GP without side information;}
\item {\textbf{LGR$^+$} -- grid-based local GP with side information.}
\end{enumerate}

All the above models implement spatiotemporal GPs defined on road networks (as described in Section \ref{sect:gp-networks}) and use the RBF kernel functions. We use a global GP (with or without side information) as the baseline for each NMF-based local GP counterpart. For each global GP, exactly $T_{\max} = 600$ observations sampled uniformly at random from $\mathbf{D}_t$ are used as its training set. We heuristically choose such value of $T_{\max}$ based on the observed tradeoff between training time and prediction error. That is, too large $T_{\max}$ would induce impractically long training time for real-time purposes, whereas too small $T_{max}$ would unacceptably increase the prediction error rate of global GPs (i.e., the under-fitting problem). The training set for each local GP consists of $\min\{T_{\max}, |S_l|\}$ observations sampled uniformly at random from the corresponding local subset $S_l$ induced by the localization of $\mathbf{D}_t$. This is to ensure fairness when comparing prediction accuracies and runtime performances between global GPs and their local counterparts.

{We also include two grid-based local GPs whose localizations are based on partitioning each city's road network into uniform spatial grids. Each local GP is learned \emph{only} from the data points belonging to a given grid cell. We then compare each grid-based local GP with its NMF-based counterpart. For fair comparisons, we set the number of grids (for the grid-based local GPs) as $K^2$, i.e., the same number of clusters used by the NMF-based local GPs.}

\begin{table}[!t]
  \centering
  \caption{GP models evaluated in the experiments. {\tt X} means `Yes'; blank means `No'.}
    \begin{tabular}{lcccc}
    \hline
    Model  & Baseline & NMF-based & Side Info & Grid-based\\
    \hline
    GP & {\tt X} & & & \\
    GP$^+$ & {\tt X} & & {\tt X} & \\
    LGP & & {\tt X} & \\
    LGP$^+$ & & {\tt X} & {\tt X} & \\
    LGR & {\tt X} & & & {\tt X} \\
    LGR$^+$ & {\tt X} & & {\tt X} & {\tt X} \\
    \hline
    \end{tabular}
  \label{tab:models}%
\end{table}%

Table \ref{tab:models} summarizes all the six models evaluated in our experiments. {All the spatial features listed in Table \ref{tab:features} are used as side information, except for longitude and latitude coordinates, which are used to define the spatiotemporal kernel function. Linear kernel functions are used for categorical variables (direction, one-way, and road type); otherwise, RBF kernels are used.}

\subsection{Evaluation Metrics and Configuration}
\label{sect:metric_setup}

We use the root mean square error (RMSE), the mean absolute error (MAE), and the mean absolute percentage error (MAPE) to evaluate the models. The three metrics are respectively defined as:
\begin{equation}
	\mathrm{RMSE} = \sqrt{\frac{1}{N} \sum_{i=1}^N (\hat{y}_i - y_i)^2},
\end{equation}
\begin{equation}
	\mathrm{MAE} = \frac{1}{N} \sum_{i=1}^N |\hat{y}_i - y_i|,
\end{equation}
\begin{equation}
	\mathrm{MAPE} = \frac{1}{N} \sum_{i=1}^N \left|\frac{\hat{y}_i - y_i}{y_i}\right|,
\end{equation}
where $\hat{y}_i$ and $y_i$ are the predicted and observed speed over road segment $i$, respectively, and $N$ is the total number of road segments in the test set. 

We also measure the runtime performances by looking at the ``wall-clock time'' (in seconds) for each model to train and make predictions at test time. This includes, whenever possible, matrix factorization, temporal cluster mapping, nearest neighbor mapping, and training and prediction time for each GP model. All our experiments were conducted on a CentOS Linux machine with $7$-core Intel(R) Xeon $2.6$ GHz processor and $70$ gigabytes of RAM.

Finally, to evaluate the significance of the improvements due to local GPs, if any, we use the non-parametric {Wilcoxon signed-rank} statistical test \cite{Wilcoxon1945}. The Wilcoxon test provides a robust alternative to the pairwise $t$-test when the measures cannot be assumed to be normally distributed.  

\subsection{Localization}
\label{sect:clustering} 

\begin{figure}[!t]
\begin{tabular}{cc}
	\includegraphics[width=1\columnwidth]{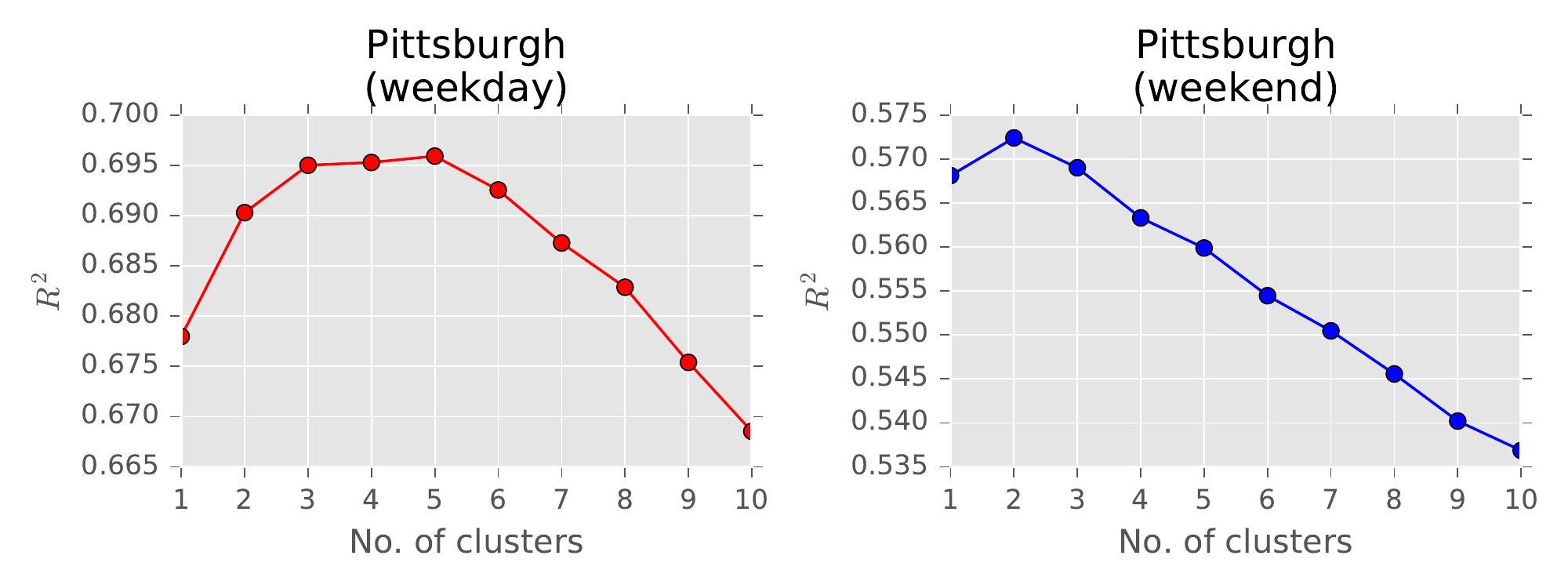}\\
	\includegraphics[width=1\columnwidth]{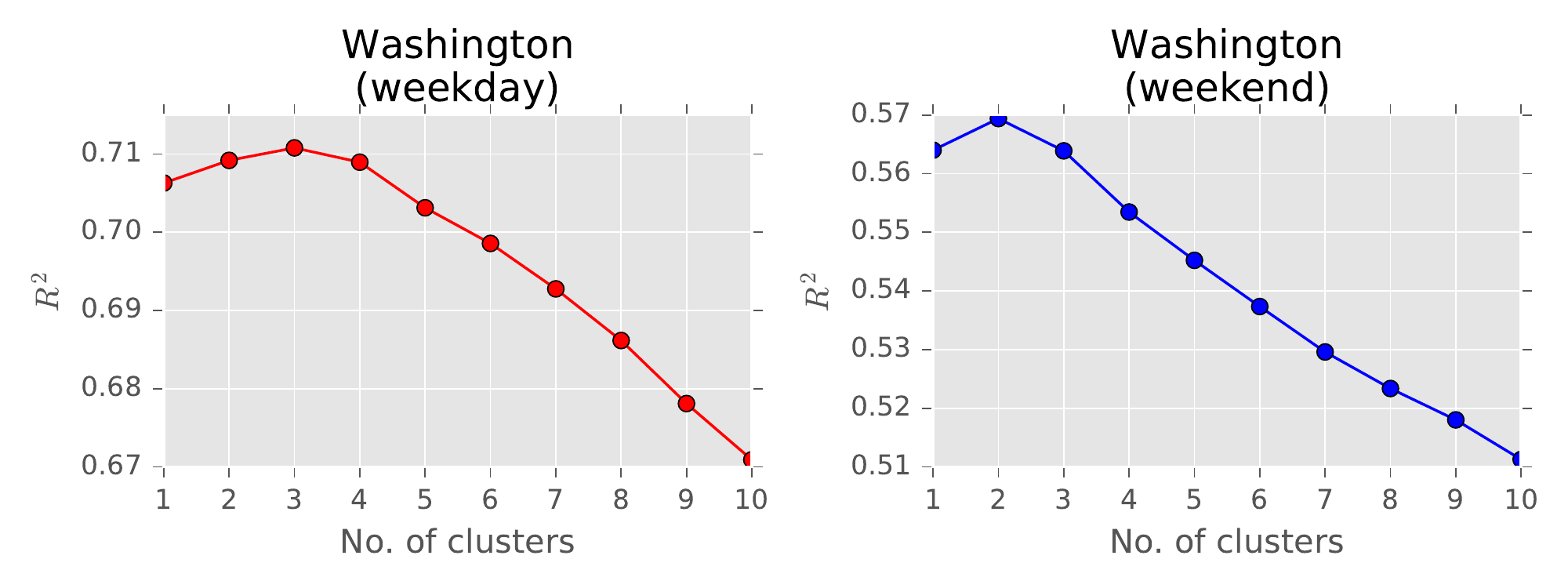}
\end{tabular}
\caption{Results of our parameter search procedure for determining $K$ for each dataset.}
\label{fig:mf-rank}
\end{figure}

{Following the procedure described in Section \ref{sect:determine_K}, we find the optimal number of clusters $K^*$ by taking $K$ that gives the highest explained variance $R^2$. To this end, we perform $10$-fold CV with $K$ varying from $1$ to $10$, and then look for the ``elbow'' point that corresponds to the highest $R^2$ averaged over $10$ folds. Fig. \ref{fig:mf-rank} shows the results. We see that the optimal $K^*$ (i.e., the ``elbow'') for Pittsburgh are $5$ for weekday and $2$ for weekend. The optimal $K^*$ for Washington are $3$ for weekday and $2$ for weekend. The higher $K^*$ for weekday suggests that the traffic patterns on the weekday are more complex than those on the weekend.}

\begin{figure}[!t]
\begin{tabular}{cc}
	\includegraphics[width=1\columnwidth]{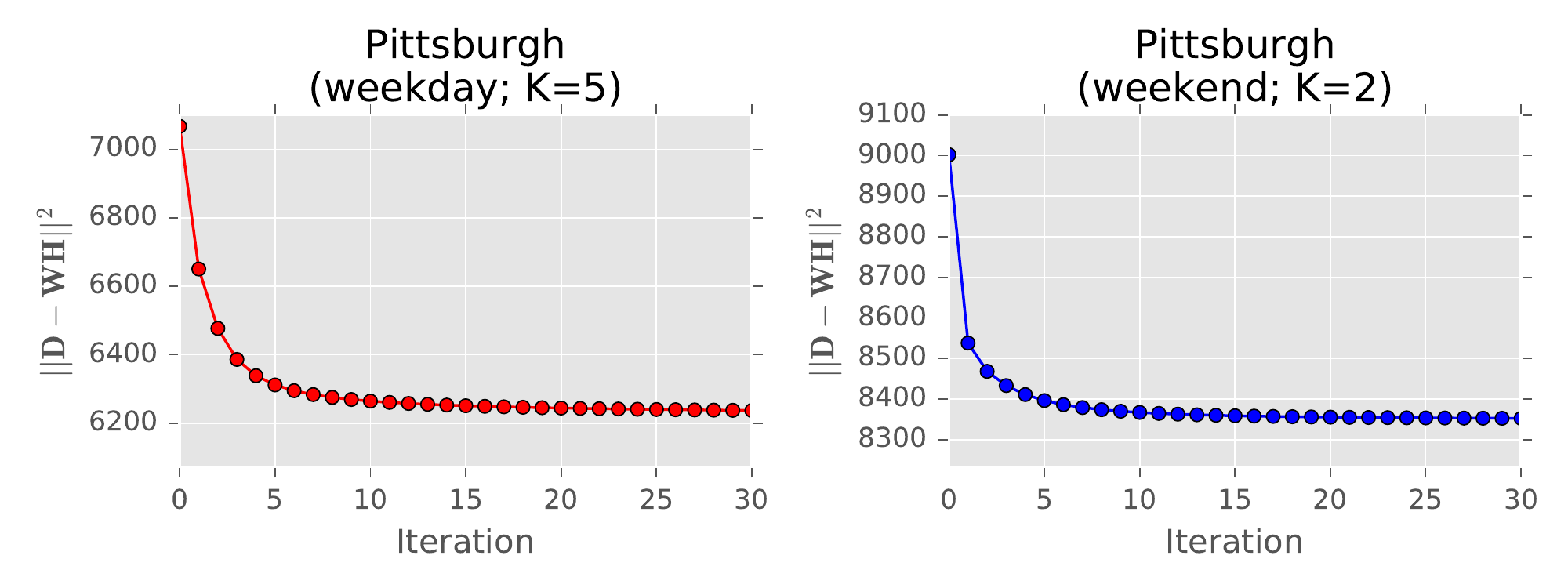}\\
	\includegraphics[width=1\columnwidth]{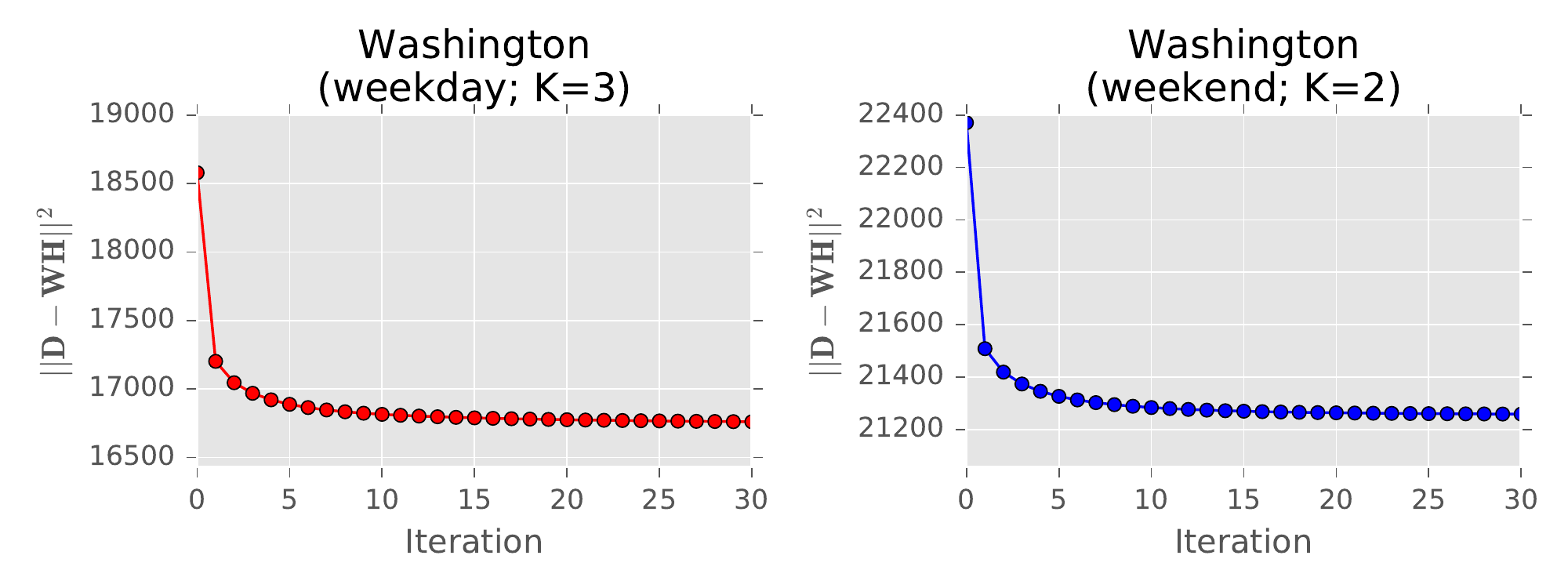}
\end{tabular}
\caption{The convergence of the coordinate-descent training in NMF using the best number of clusters $K^*$ for each dataset.}
\label{fig:mf-convergence}
\end{figure}

{
To verify the convergence of the CD algorithm, we also monitor the residual error $||\mathbf{D} - \mathbf{W} \mathbf{H}||^2$ (i.e., the first term in equation (\ref{eqn:sparse_nmf_loss})) over different training iterations. Fig. \ref{fig:mf-convergence} shows the convergence plots of $||\mathbf{D} - \mathbf{W} \mathbf{H}||^2$ for different datasets. Here, we zoom into the first $30$ training iterations (out of a total of $200$ iterations as per Section \ref{sect:nmf_cd}) in order to see more clearly the convergence of $||\mathbf{D} - \mathbf{W} \mathbf{H}||^2$. Indeed, $||\mathbf{D} - \mathbf{W} \mathbf{H}||^2$ converges rapidly within $10$ iterations and no longer decreases substantially afterwards. This shows that the CD algorithm offers an efficient method for training NMF.
}

\begin{figure}[!t]
\centering
\subfloat[] {
	\includegraphics[width=0.40\textwidth]{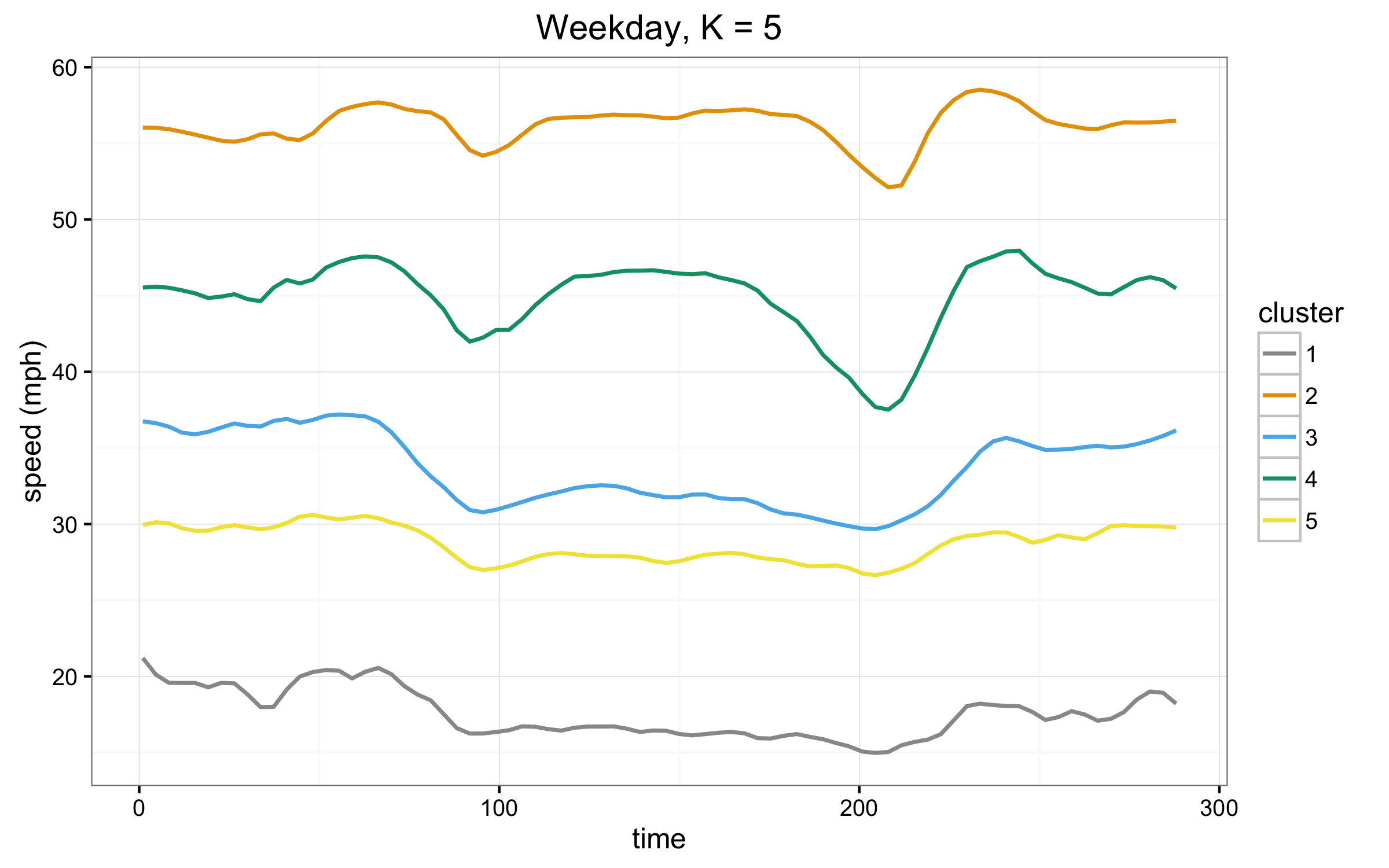}
	\label{fig:spatial-clusters}
} \par
\subfloat[] {
	\includegraphics[width=0.38\textwidth]{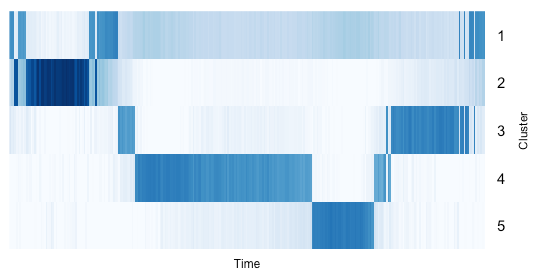}
	\label{fig:temporal-clusters}
}
\caption{(a) Time series of the average speed along clustered road segments for weekday data in Pittsburgh. Horizontal axis shows the $5$-minute intervals. (b) Heat map of the column-wise normalized matrix $\mathbf{H}$ that visualizes the temporal clustering for weekday data in Pittsburgh. Bolder shade means closer to $1$ and lighter means closer to $0$.}
\label{fig:cl-time-series}
\end{figure}

Fig. \ref{fig:spatial-clusters} illustrates the time series of the average speed along the clusters of road segments every $5$-minute interval on a typical weekday in Pittsburgh. Our NMF method has clustered the road segments into different types, each having different throughput and daily speed distribution. For example, for clusters $2$, $3$, and $4$, we can see clearly the rush hour effects observed earlier in Fig. \ref{fig:time-series} to different levels. These clusters mostly contain road segments leading to (and away from) the business areas in the downtown. The other clusters with slower speeds contain mostly small segments in the residential areas, or those that are in the business areas but do not lead to the residential areas.

Fig. \ref{fig:temporal-clusters} presents a heatmap visualization of the temporal cluster mapping derived from the column-wise normalized matrix $\mathbf{H}$ on a typical weekday in Pittsburgh. The result shows clear temporal patterns of the traffic speed in the city, whereby the probabilistic assignment of the temporal clusters is sparse. That is, at a given time step, only a few clusters (darker shades) have substantially higher probability value  than the rest (lighter shades). In this case, we can identify rush hours by looking at rapidly changing cluster assignments that occur within a fairly short period of time.

It is worth noting that because of the temporal sparsity problem mentioned in Section \ref{sect:dataset}, each $\mathbf{D}_t$ of each dataset has a significant number of missing values. 
NMF solves this problem by optimally imputing those missing values while imposing non-negativity and sparsity constraints.

\subsection{Evaluation Results}
\label{sect:results}

Table \ref{tab:nmf_runtime} shows the summary statistics of the NMF-based localization runtime. We can see that, on average, NMF-based localization is sufficiently fast for most real-time applications (much less than $1$ second) for all datasets.

\begin{table}[!t]
\centering
\caption{Runtime statistics (in seconds) of NMF-based localization for Pittsburgh (PGH) and Washington (WAS) on weekday (WD) and weekend (WE).}
\begin{tabular}{lrrr}
\hline
City (day) &      Mean &   Median &       Stdev \\
\hline
PGH (WD) &  0.3137 &  0.2676 &  0.1243 \\
PGH (WE) &  0.1889 &  0.1621 &  0.1043 \\
WAS (WD) & 0.2597 & 0.2050 & 0.1071 \\
WAS (WE) & 0.1395 & 0.1146 & 0.0490 \\
\hline
\end{tabular}
\label{tab:nmf_runtime}
\end{table}

\begin{figure*}[!t]
\begin{center}
	\includegraphics[width=0.75\textwidth]{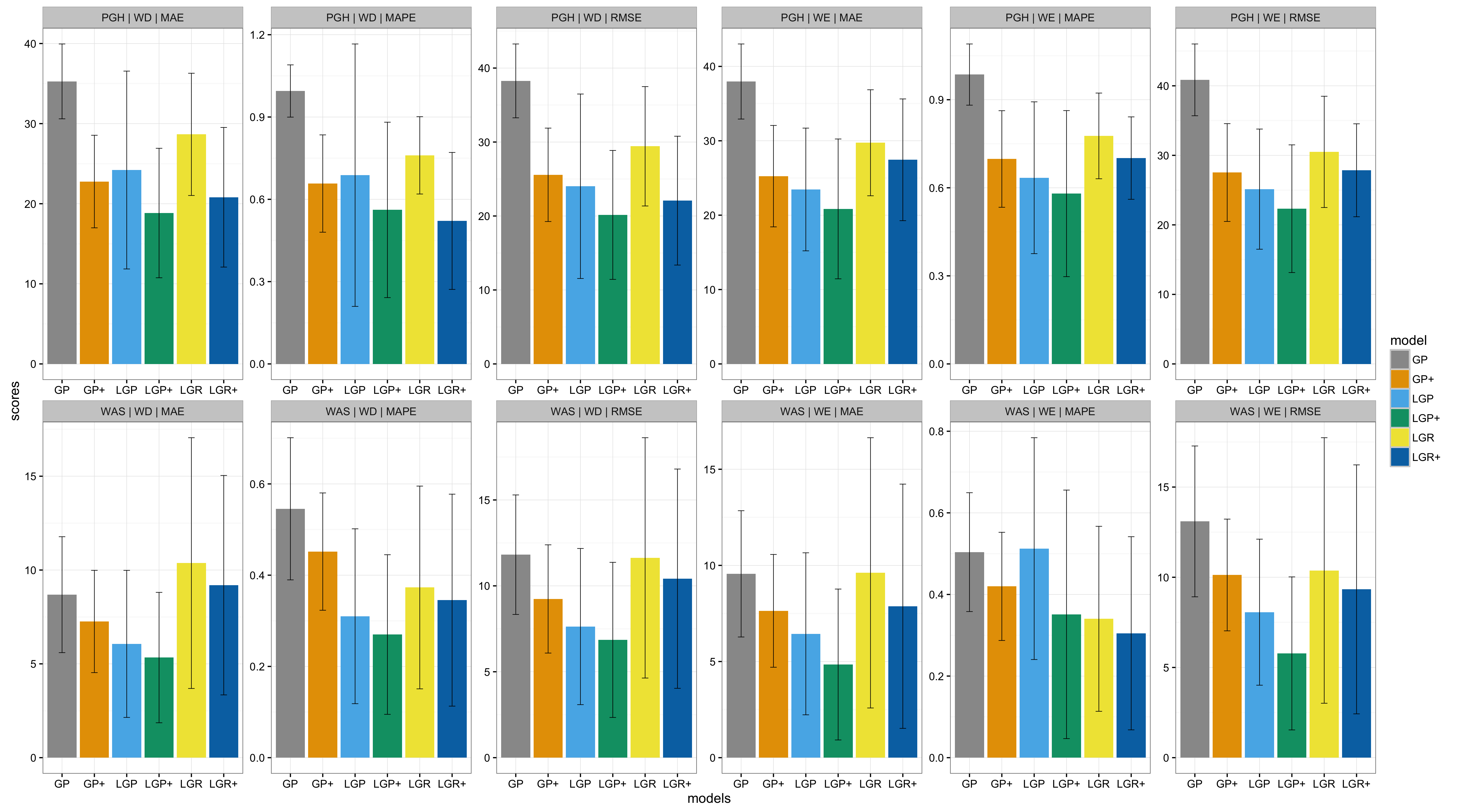}
\end{center}
\vspace{-2ex}
\caption{Evaluation of speed prediction across the six models using the metrics: MAE, MAPE, and RMSE. Datasets evaluated are: Pittsburgh (PGH) and Washington (WAS) on weekday (WD) and weekend (WE).}
\label{fig:all-scores}
\vspace{0ex}
\end{figure*}

\begin{figure}[!t]
\begin{center}
	\includegraphics[width=0.45\textwidth]{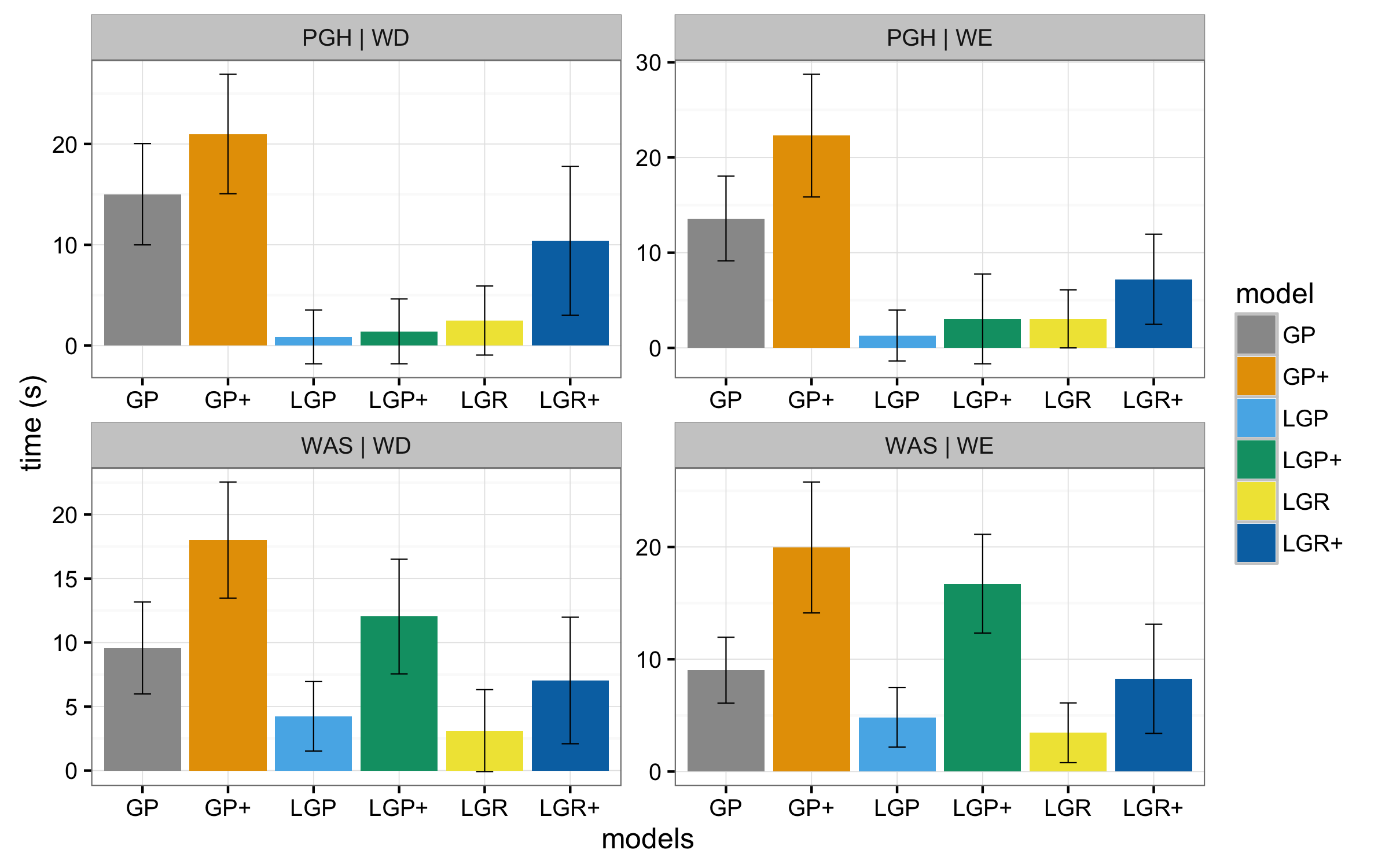}
\end{center}
\vspace{-2ex}
\caption{Evaluation of the total runtime performances across six models for the two cities of Pittsburgh (PGH) and Washington (WAS) on weekday (WD) and weekend (WE).}
\label{fig:all-times}
\vspace{0ex}
\end{figure}

{Fig. \ref{fig:all-scores} shows the prediction evaluation results of the six GP models listed in Table \ref{tab:models} for both Pittsburgh (PGH) and Washington (WAS) across the three evaluation metrics (MAE, MAPE, and RMSE) averaged over all the trials on both test weekday (WD) and weekend (WE). For Pittsburgh (top row), it can be seen that global GPs without side information always have the highest error rates. Grid-based local GPs perform better than global GPs; however, the predictions with the lowest errors come from NMF-based local GPs. Having side information always improve prediction accuracies with weekdays having stronger effects than weekends. Side information has the strongest effects on global GPs, which is not surprising given its largely diffuse training set. All the three metrics display consistent observations with MAPE having the highest variance. Our pairwise Wilcoxon tests between global GPs and NMF-based local GPs (with/without side information) and between NMF-based local GPs and grid-based local GPs (with/without side information) are all significant at the $5\%$ level, except for LGP$^+$ and LGR$^+$ for weekday data evaluated using MAPE. It can be argued that NMF-based local GPs with side information is the best-performing model overall. This demonstrates the effects of learning from a smaller, but \textit{more relevant} local subsets of training data \cite{snelson2007local}.}

{For Washington, similar observations can be seen in Fig \ref{fig:all-scores} (bottom row). Global GPs without side information almost always have the highest error rates. Grid-based local GPs yield high variances and, at the same time, perform much worse than those in Pittsburgh (when compared to global GPs). This showcases the inability of simple spatial grid partitioning to adequately model more complex traffic patterns in a completely different urban setting. Having side information invariably reduces error rates for all the models. Similar pairwise Wilcoxon tests were performed, all of which are significant at the $5\%$ level, except for the three pairs: GP vs. LGP, GP$^+$ vs. LGP$^+$, and LGP$^+$ vs. LGR$^+$ for weekend data evaluated using MAPE due to high variances. It can thus be concluded that NMF-based local GPs with side information is the best-performing model for weekday data. It is, however, inconclusive for weekend data.} 

{Fig. \ref{fig:all-times} shows the evaluation of runtime performances for all the models. For Pittsburgh (top row), NMF-based local GPs significantly outperform global GPs by more than $10$ folds (i.e., NMF-based local GPs are more than $10$ times faster) for weekday, with and without side information. Higher $K^*$ significantly reduces the runtime of local GPs as evidenced by shorter runtime on the weekday compared to that on the weekend. Apart from that, we see a similar pattern for weekend: both local GPs significantly outperform global GPs in terms of runtime, and NMF-based local GPs are more than $6$ times faster. Having side information invariably improves prediction accuracies, but also increases runtime for all models. This is particularly true for grid-based local GPs, which suggests that the chosen set of side information induces more complex correlation structure (hence, parameter estimates) for GP learning . All Wilcoxon pairwise tests are statistically significant at the $5\%$ level.}

{For Washington, Fig. \ref{fig:all-times} (bottom row) shows similar observations: local GPs are faster than global GPs and having side information increases runtimes. What is interesting, however, is the observations that NMF-based local GPs with side information have significantly higher runtimes than grid-based local GPs. This might be due to the need of LGP$^+$to model more complex local subsets that results from non-uniform partitioning of training data than LGR$^+$. We further discuss this observation in the following section. All Wilcoxon pairwise tests are significant at the $5\%$ level.}

\subsection{Discussion}
\label{sect:discussion}

\begin{figure*}[!t]
\begin{center}
	\includegraphics[width=0.80\textwidth]{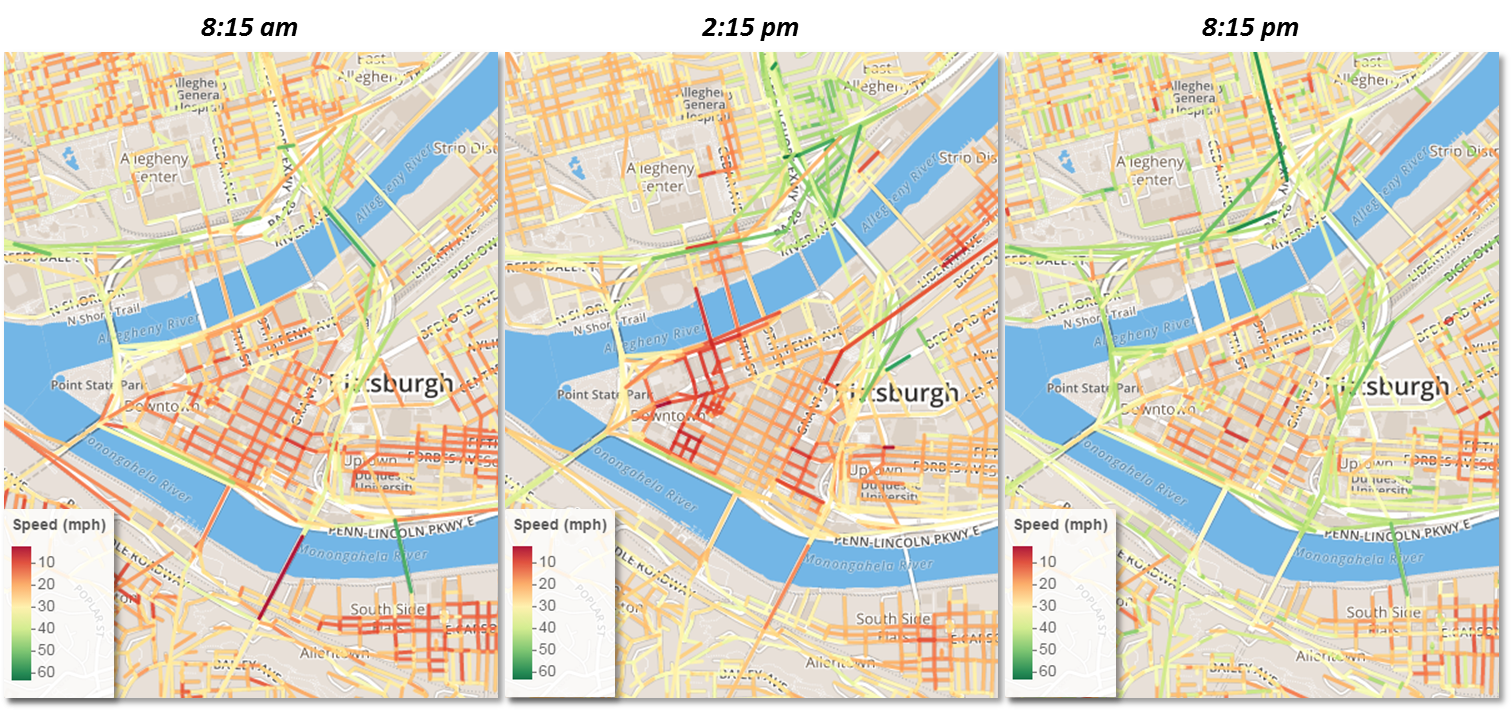}
\end{center}
\vspace{-2ex}
\caption{Visualization of the spatiotemporal inference of traffic speed in Pittsburgh on the test {weekday} (August 28, 2014) using the LGP$^+$ model. Training data obtained using the sliding window method from $5\%$ of the road network and at three time points: $8$ a.m., $2$ p.m., and $8$ p.m. Test time for each is a three-step ahead prediction (i.e., in $15$ minutes).}
\label{fig:weekday-speed}
\vspace{0ex}
\end{figure*}

For all datasets, global GPs incur high runtimes and have low prediction accuracies, which render them impractical for real-time applications. Local GPs thus become viable solutions to real-time traffic prediction with significantly lower runtime costs, with and without side information. Local GPs with side information can give more accurate predictions but at increased time costs, and thus are more suitable for longer-horizon applications. On the other hand, local GPs without side information are more suitable for shorter-horizon applications, where decisions are to be made fast.

{In most cases, NMF-based local GPs predict significantly better than grid-based local GPs, as shown in Fig. \ref{fig:all-scores}. We have also seen that, for the same set of side information features, different localization methods can result in significantly different runtimes for training local GPs. This is due to our uniform (and uninformed) selection of the same set of side information listed in Table \ref{tab:features} for both cities. Different cities induce different traffic phenomena and optimization problems (and complexities). It is unreasonable that the same set of side information is able to model those distinct phenomena equally effectively and efficiently. Discriminatory feature selection should have been exercised. Feature selection is an entire different issue and often relies on domain knowledge; thus, it is out of scope of this paper.}

In practice, one needs to trade off between model expressiveness (i.e., incorporating side information) and efficiency depending on one's sensitivity to accuracy and time. How to select side information also matters. In this respect, it is important to consider the most relevant side information (and the smallest subset of such) to the traffic phenomenon being modeled in order to maximize its benefits. Such knowledge also belongs to the domain expert.



Finally, Fig. \ref{fig:weekday-speed} visualizes the spatiotemporal inferences of traffic speed on \textit{the entire} road network of Pittsburgh (zoomed into the downtown area) on the test weekday at three test times: $8$:$15$ a.m., $2$:$15$ p.m., and $8$:$15$ p.m. NMF-based local GPs with side information were used to make the inferences. The training sets were derived using the sliding window method at time $t \in \{8, 14, 20\}$ hours. Each test time is a $3$-step ahead prediction. At each test time, the observed speeds cover $5\%$ of the whole network (while prediction makes for the entire of it). Fig. \ref{fig:weekday-speed} shows clearly the morning rush hour effect at $8$:$15$ a.m., where the main roads leading to the downtown and other business areas become highly congested (with lower speed distribution). At $2$:$15$ p.m., congestion becomes more localized to the business areas because of office hours, while the main roads have become visibly more cleared of traffic. At $8$:$15$ p.m., traffic on the whole gets visibly faster with main roads leading to and from the business areas having apparently much faster flows, and congested areas have now become more localized to the nightlife areas in downtown.

\section{Conclusion}
\label{sect:conclusion}

This paper addresses an important problem in urban computing: real-time traffic speed modeling and prediction. To this end, we propose the novel idea of localizing spatiotemporal Gaussian processes (GPs) using non-negative matrix factorization (NMF). In addition, we make use of the expressiveness of GP kernel functions to model traffic speed through directed links of a road network and incorporate side information via additive kernel. Extensive empirical studies using real-world traffic data collected in diverse geospatial settings have demonstrated the efficacy of our proposed approach, in terms of both computational efficiency and prediction accuracy, against the baseline global and local GPs. We also show that a tradeoff exists between model expressiveness and runtime performance when side information is taken into account. It is therefore important to consider the most relevant side information for that matter. 

\ifCLASSOPTIONcompsoc
\section*{Acknowledgments}
\else
\section*{Acknowledgment}
\fi
{This research is supported by the National Research Foundation, Prime Minister's Office, Singapore under its International Research Centres in Singapore Funding Initiative.

Siyuan Liu is additionally supported by the Basic Research Program of Shenzhen: JCYJ20140610152828686, and the Natural Science Foundation of China: 61572488.

We would also like to thank Sean Qian of Carnegie Mellon University for valuable discussions during this work.
}

\ifCLASSOPTIONcaptionsoff
  \newpage
\fi

\small
\bibliography{bibliography}  
\bibliographystyle{IEEEtran}  

\vspace{-5ex}

\begin{IEEEbiography}[{\includegraphics[width=1in,height=1.25in,clip,keepaspectratio]{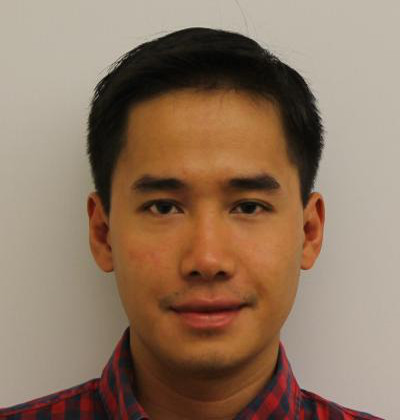}}]{Truc Viet Le}
is a Ph.D. candidate in Information Systems at the Singapore Management University. His research interests include methods for modeling and predicting urban mobility patterns and developing smart urban transportation technologies. He spent the year of 2015 at Carnegie Mellon University in Pittsburgh, P.A. during his Ph.D. tenure. He obtained his Bachelor's and Master's degree from Nanyang Technological University in 2009 and 2012, respectively.
\end{IEEEbiography}

\vspace{-5ex}

\begin{IEEEbiography}[{\includegraphics[width=1in,height=1.25in,clip,keepaspectratio]{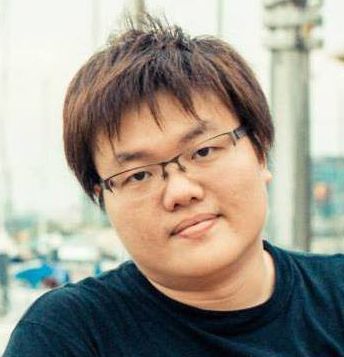}}]{Richard Oentaryo}
is a Research Scientist at the Living Analytics Research Centre, Singapore Management University. He obtained his Ph.D. and B.Eng. (First Class) in Computer Engineering from the Nanyang Technological University, Singapore, in 2011 and 2004 respectively. His research interests include machine learning, data mining, and nature-inspired computing. Dr. Oentaryo has published more than 25 papers in various international journals and conferences.
\end{IEEEbiography}

\vspace{-5ex}

\begin{IEEEbiography}[{\includegraphics[width=1in,height=1.25in,clip,keepaspectratio]{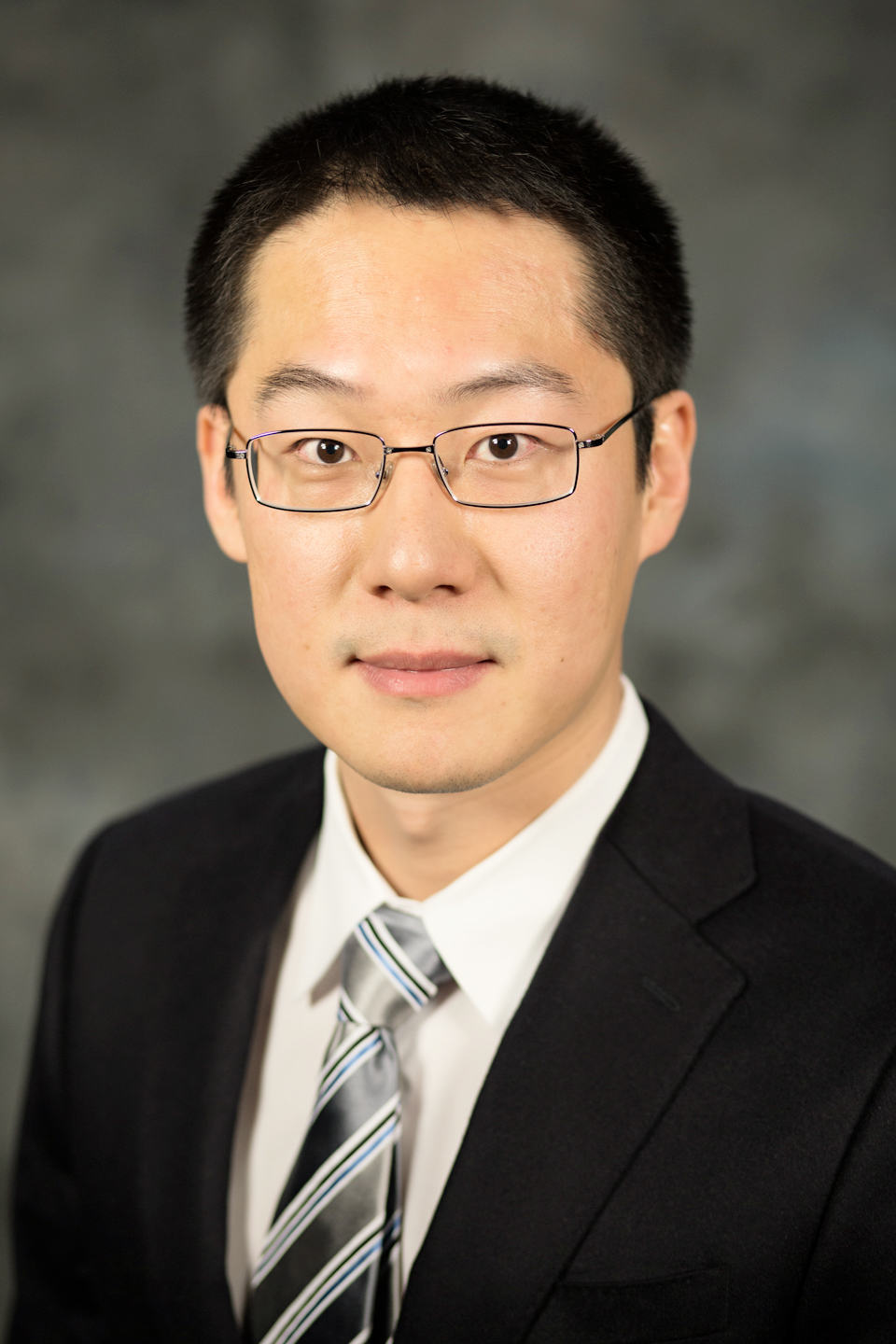}}]{Siyuan Liu}
is an Assistant Professor at the Smeal College of Business, Pennsylvania State University. He received his first Ph.D. degree from the Department of Computer Science and Engineering at Hong Kong University of Science and Technology, and his second Ph.D. from the University of Chinese Academy of Sciences. His current research interests include spatial and temporal data mining, social networks analytics, and mobile marketing.
\end{IEEEbiography}

\vspace{-5ex}

\begin{IEEEbiography}[{\includegraphics[width=1in,height=1.25in,clip,keepaspectratio]{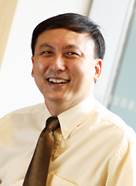}}]{Hoong Chuin Lau}
is a Professor of Information Systems and Director of the Fujitsu-SMU Urban Computing and Engineering Corp Lab at the Singapore Management University. Prior to that, he was a research scientist at the Institute of Infocomm Research in Singapore (1997--1999), and Assistant Professor at the School of Computing, National University of Singapore (2000--2005). He obtained his Doctorate of Engineering degree in Computer Science from the Tokyo Institute of Technology in 1996, and B.Sc. and M.Sc. degrees in Computer Science from the University of Minnesota. 
\end{IEEEbiography}

\end{document}